\documentclass[10pt,twocolumn,letterpaper]{article}

\usepackage{iccv}              

\usepackage{graphicx}
\usepackage{amsmath}
\usepackage{amssymb}
\usepackage{caption}
\usepackage{subcaption}
\usepackage{float}
\usepackage{mathptmx}
\usepackage{algorithm}
\usepackage{algpseudocode}
\usepackage{amsfonts}
\usepackage{booktabs}
\usepackage{colortbl}
\usepackage{xcolor}
\usepackage{multirow}
\usepackage{makecell}
\usepackage{tabularx}
\usepackage{tcolorbox}

\newtheorem{theorem}{Theorem}[section]

\newtheorem{prop}{Proposition}[section]
\newtheorem{definition}{Definition}[section]

\definecolor{iccvblue}{rgb}{0.21,0.49,0.74}
\usepackage[pagebackref,breaklinks,colorlinks,allcolors=iccvblue]{hyperref}

\def\paperID{9461} 
\def\confName{ICCV}
\def\confYear{2025}

\title{Decomposing Task Vectors for Refined Model Editing}

\author{%
  \begin{tabular}{c}
    \textbf{Hamed Damirchi$^{1}$ \qquad Ehsan Abbasnejad$^{2}$} \qquad
    \textbf{Zhen Zhang$^{1}$ \qquad Javen Shi$^{1}$} \\[1ex]
    \textnormal{$^{1}$Australian Institute for Machine Learning, Adelaide University \qquad
               $^{2}$Monash University} \\[0.5ex]
    \small{\texttt{$^{1}$\{firstname.lastname\}@adelaide.edu.au \quad $^{2}$\{firstname.lastname\}@monash.edu}}
  \end{tabular}
}

\begin{document}
\maketitle
\begin{abstract}
Large pre-trained models have transformed machine learning, yet adapting these models effectively to exhibit precise, concept-specific behaviors remains a significant challenge. Task vectors, defined as the difference between fine-tuned and pre-trained model parameters, provide a mechanism for steering neural networks toward desired behaviors. This has given rise to large repositories dedicated to task vectors tailored for specific behaviors.
The arithmetic operation of these task vectors allows for the seamless combination of desired behaviors without the need for large datasets. 
However, these vectors often contain overlapping concepts that can interfere with each other during arithmetic operations, leading to unpredictable outcomes. We propose a principled decomposition method that separates each task vector into two components: one capturing shared knowledge across multiple task vectors, and another isolating information unique to each specific task. By identifying invariant subspaces across projections, our approach enables more precise control over concept manipulation without unintended amplification or diminution of other behaviors. We demonstrate the effectiveness of our decomposition method across three domains: improving multi-task merging in image classification by 5\% using shared components as additional task vectors, enabling clean style mixing in diffusion models without generation degradation by mixing only the unique components, and achieving 47\% toxicity reduction in language models while preserving performance on general knowledge tasks by negating the toxic information isolated to the unique component. Our approach provides a new framework for understanding and controlling task vector arithmetic, addressing fundamental limitations in model editing operations.
\end{abstract}    
\section{Introduction}
\label{sec:intro}

Task vectors \cite{task_vec} are directions in the parameter space defined by the difference between fine-tuned and pre-trained model weights. These vectors encapsulate the information necessary for a pre-trained model to perform well on a specific task, enabling model editing~\cite{task_vec}, concept negation~\cite{ortizjimenez2023task}, efficient fine-tuning~\cite{Yang2023AdaMergingAM} and test-time adaptation~\cite{zhang2024atlas} with a negligible number of parameters. Arithmetic operations with these vectors enable conceptual compositions to achieve desired behaviors without extensive retraining. This approach presents a promising strategy for adapting pre-trained models to a wide range of domains and tasks. This has given rise to an extensive repositories of task vectors, particularly in their low-rank form obtained through LoRAs~\cite{hu2022lora}.

\begin{figure*}[t]
    \centering
    \includegraphics[width=0.97\textwidth]{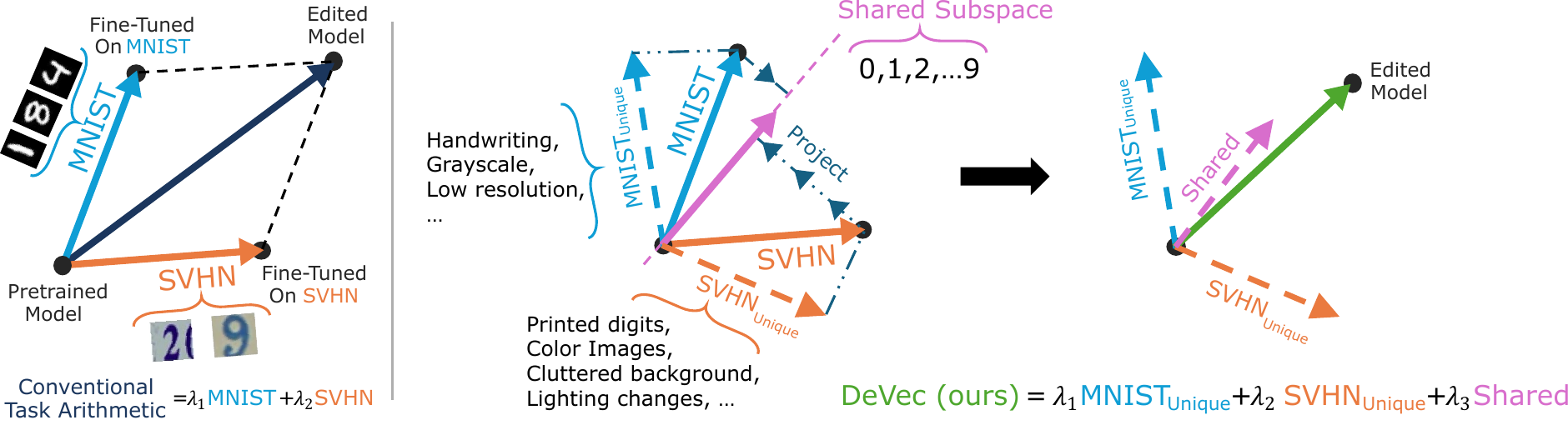}
    \caption{Unlike conventional task arithmetic approaches, we start by \textbf{(1)} Finding the subspaces \textcolor[HTML]{D86ECC}{shared} between available task vectors. \textbf{(2)} \textcolor[HTML]{156082}{Project} each task vector onto the shared subspaces to derive the \textcolor[HTML]{D86ECC}{shared} component. \textbf{(3)} Subtract the shared component from the original task vector to derive the unique component (dashed vectors). \textbf{(4)} Recompose the decomposed components with component-specific coefficients $\lambda_i$, allowing for more controlled model editing.}
    
    \label{fig:figure1}
    \vspace{-5mm}
\end{figure*}

While task vectors capture dataset-specific directions in the parameter space,
in practice, it is nearly impossible to find a dataset that shares no conceptual overlap with the pre-trained model or other datasets. Consequently, when performing arithmetic operations on these task vectors, such conceptual overlaps can introduce interferences, potentially causing certain concepts to be overemphasized or underemphasized, resulting in unpredictable outcomes. 
In a simple digit recognition task, a task vector trained on MNIST~\cite{lecun2010mnist} and another trained on SVHN~\cite{svhn} reveal shared patterns for recognizing digits, while also capturing distinct features influenced by differences in background, style, and color (See \cref{fig:figure1}).
Therefore, to mitigate such interference, it is crucial to consider both the commonalities and the unique characteristics of task vectors when performing arithmetic operations. Moreover, accounting for these factors enables a more controlled composition of concepts, allowing for precise model editing tailored to downstream tasks.

In this paper, we propose a method to decompose each task vector into two distinct components using column space projections of weight matrices. One component captures shared knowledge across the available task vectors, while the other isolates information unique to each task.
We accomplish this by identifying subspaces where task vectors overlap while isolating the remaining subspaces as unique components. Specifically, we identify directions in the parameter space that remain largely unchanged when projected by all task vectors, designating them as shared.
Each task vector is then projected onto this shared subspace, and its unique component is obtained by subtracting the shared part. This decomposition provides finer control over subspaces that were previously entangled within each task vector, allowing manipulation without inadvertently amplifying or diminishing other components. This manipulation is performed in parameter space by linearly adjusting the pretrained model’s weight matrices toward the desired concepts embedded in each component. 

While enabling more precise manipulation of concepts within task vectors, the use of each component (shared or unique) depends on the application. We demonstrate our method in image classification, style mixing for image generation, and toxicity reduction in large language models. In image classification, decomposing task vectors improves multi-task merging and generalization for unseen tasks by providing finer concept control through separate decomposed component coefficients in the linear adjustment process. For style mixing, linearly combining multiple style task vectors can degrade image generation quality, whereas mixing only their unique components preserves quality regardless of the number of styles. In the case of language models, reducing toxicity without decomposition lowers performance on general knowledge and reasoning tasks. By isolating the unique component of toxic task vectors for manipulation using this component, we reduce toxicity on two benchmarks while maintaining overall performance. We further analyze the decomposition process through ablation studies, providing deeper insights into its effects on each application. Our main contributions are as follows:
\begin{itemize}
    \item We propose a principled approach for decomposing task vectors into shared and unique components through column space projections. This enables precise isolation of task-specific behaviors from common subspaces across multiple tasks.
    
    \item Our decomposition method provides a new framework for understanding and controlling task vector arithmetic, addressing fundamental limitations in model editing operations by preventing concept interference and undesired behavioral transfers between task vectors.
    
    \item Through comprehensive experiments across three domains - image classification, image generation, and language models - we demonstrate that our approach enables more effective model editing, achieving state-of-the-art results in toxicity removal (47\% less toxicity on the ToxiGen benchmark), improving multi-task learning by 5\%, and enabling clean style mixing in diffusion models without generation degradation. Our ablation studies provide insights into the relationship between shared subspaces and task-specific behaviors in neural networks.
\end{itemize}
\section{Related Work}
\label{sec:relatedwork}
Model editing methods fall into three main paradigms: task arithmetic, model merging, and model surgery. Task arithmetic manipulates parameter or activation-space directions representing desired behaviors, enabling controlled model edits while preserving general capabilities. We focus specifically on parameter-space task arithmetic. Model merging interpolates parameters from multiple fine-tuned models to retain characteristics from each. In contrast, model surgery takes a more targeted approach, manipulating specific parameters or features within the model architecture. In this section, we review recent developments in each category and position our approach within this landscape.

Task arithmetic \cite{task_vec} uses the difference between the fine-tuned and pre-trained parameters of a model as the direction representing the dataset used for obtaining the fine-tuned parameters. The set of obtained parameters for each layer are then treated as direction in the parameter space, named task vector, that can be added to or subtracted from the pre-trained model to edit model behavior. \cite{ortizjimenez2023task} then takes the fine-tuning process to the model's tangent space to promote weight disentanglement and allow for control over localized regions in the weight space. AdaMerging \cite{Yang2023AdaMergingAM} and aTLAS \cite{zhang2024atlas} propose methods to learn the coefficient of multiple task vectors in a scenario where the model and vectors are both frozen and the only learnable parameters are the coefficients, leading to model editing with a significantly smaller number of trainable parameters. Our method is complementary to any training approach for task vectors. By decomposing the available vectors into unique and shared components, we allow for finer control over the editable concepts the vectors provide. 

Model merging approaches have gained attention for their ability to combine multiple fine-tuned models without additional training. Early works demonstrated that simply averaging the weights of models fine-tuned from the same initialization can preserve their individual capabilities \cite{wortsman2021robust}. Fisher merging \cite{fishermerging} improved upon this by using the diagonal of each model's Fisher Information Matrix as weights when averaging parameters, providing better preservation of important parameters for each task. RegMean \cite{jin2023dataless} proposed solving a least-squares regression problem to minimize the distance between the merged model's activations and individual models' activations demonstrating that model merging can still be effective without access to the original training data, making it particularly valuable when data privacy is a concern. TIES-merging \cite{yadav2023tiesmerging} addresses interference between parameters of different models by first trimming redundant parameters and resolving sign conflicts before merging, leading to better preservation of task-specific capabilities in the merged model. In this work, we aim to isolate the shared from the unique knowledge between the available fine-tuned models and gain control over the manipulable characteristics that each model possesses as opposed to interpolation of models which may be suboptimal for model editing purposes due to a lack of fine-grained control over specific concepts.

Narrowing the focus onto specific parameters, Model Surgery \cite{wang2024modelsurgerymodulatingllms} uses linear probing to identify rows of the weight matrices of a large language model that are responsible for certain behaviors such as toxicity and susceptibility to jailbreaks (prompting the model in a way to avoid adherence to the given safety instructions). This is done by training a linear probe on the desired behavior for each layer of the model and then linearly adding or removing the weights from the pre-trained model. We do not require access to the activations of the model and we only perform simple projection and linear operations within the parameter space. Additionally, task vectors allow for addition or removal of information on top of enabling model editing for achieving desired behavior. This added information is naturally severely limited when narrowing the focus to a fraction of the parameters of the model.
\section{Methodology}
\label{sec:formatting}

A task vector is defined as the difference between the pre-trained and fine-tuned parameters at each layer of a model $M$, parameterized by a set of weights $W$ and biases $b$. In particular, for a given task $t$, the task vector $\textbf{W}_t^l$ at layer $l$ is computed as $\textbf{W}_t^l = \textbf{W}_{\text{ft}}^l - \textbf{W}_{\text{pre}}^l$, where $\textbf{W}_{\text{pre}}^l$ and $\textbf{W}_{\text{ft}}^l$ represent the pre-trained and fine-tuned weights, respectively. The available task vectors can then linearly modify the model with per-vector coefficients $\lambda_t$ similar to $\textbf{W}^l = \textbf{W}_{\text{pre}}^l + \sum_t\lambda_t*\textbf{W}_t^l$. 
For brevity, we drop the explicit dependence on layer $l$.
We consider the problem of decomposing a task vector into shared and unique components prior to this model editing operation. We define the shared component as the subspaces on which a task vector operates that happen to be shared between multiple task vectors. The concept of "shared knowledge" is hence defined here with respect to a pool of available task vectors. This leaves the unique component to be the particular subspaces that would only be modified through task vectors specific to each dataset. This way, our approach extracts shared components by identifying directions in parameter space where all task vectors exhibit significant transformations, distinguishing them from directions uniquely important to individual task vectors.

\subsection{Identifying Shared Subspaces in Task Vectors}
Without loss of generality, we begin by focusing on two task vectors in this section, with the extension to multiple vectors discussed thereafter. 
As mentioned, we start by identifying shared directions. To do this, we project one task vector onto the column space of another. This process hinges on the fact that subspaces not shared between the two task vectors will be altered, leading to corresponding eigenvalues significantly smaller than 1 when projected through both vectors. This is while shared subspaces would remain largely unchanged with eigenvalues close to 1. We first define projection matrices before detailing the approach.

\begin{definition}[Projection Matrix \mbox{\cite{golub_matrix_2013}}] 
Let $\mathbf{W} \in \mathbb{R}^{n \times m}$ be a weight matrix defining a task vector at a layer of model $M$ with singular value decomposition $\mathbf{W}=\mathbf{U}\Sigma\mathbf{V}^T$, where $\mathbf{U}$, $\mathbf{V}$, and $\Sigma$ are the left and right singular vectors and the singular values, respectively.  The projection matrix $\mathbf{P}$ onto the column space of $\mathbf{W}$ is defined as:
\begin{align}
    \mathbf{P} &= \mathbf{W}(\mathbf{W}^T\mathbf{W})^{-1}\mathbf{W}^T \\
    &= \mathbf{U} \Sigma \mathbf{V}^T (\mathbf{V} \Sigma^T \mathbf{U}^T \mathbf{U} \Sigma \mathbf{V}^T)^{-1} \mathbf{V} \Sigma \mathbf{U}^T \\
      &= \mathbf{U} \Sigma (\Sigma^T \Sigma)^{-1} \Sigma^T \mathbf{U}^T \\
      &= \mathbf{U} \mathbf{U}^T.
\end{align}
\end{definition}
Given that the left singular vectors of $\mathbf{W}$ (columns of $\mathbf{U}$) are also eigenvectors of $\mathbf{WW}^T$ and that $\mathbf{P}$ projects onto the column space of $\mathbf{W}$, we have $\mathbf{P}\mathbf{z}=\mathbf{z}$ where $\mathbf{z}$ is an eigenvector of $\mathbf{W}$. This property shows that after projection, directions inline with the column space of the task vector will remain unaltered. Using this property, we define our method for task vector decomposition.

\begin{prop}[Task Vector Decomposition]
\label{theorem}
Let $\mathbf{W}_1, \mathbf{W}_2 \in \mathbb{R}^{n \times m}$ represent the weight matrices associated with the respective task vectors. We consider $\mathbf{W}_1=\mathbf{U}_1 \Sigma_1 {\mathbf{V}_1}^T$, $\mathbf{W}_2=\mathbf{U}_2 \Sigma_2 {\mathbf{V}_2}^T$ obtained from their singular value decomposition (SVD).
Let $\mathbf{P}_1 = \mathbf{U}_1\mathbf{U}_1^T$ and $\mathbf{P}_2 = \mathbf{U}_2\mathbf{U}_2^T$ be their respective projection matrices onto their column spaces. The eigenvectors of $\mathbf{P}_{1,2} = \mathbf{P}_1\mathbf{P}_2$ corresponding to eigenvalues of 1 span the shared subspace between $\mathbf{W}_1$ and $\mathbf{W}_2$.
\end{prop}

\emph{Proof}
$\mathbf{P}_1$ and $\mathbf{P}_2$ are projections onto the columns space of $\textbf{W}_1$ and $\textbf{W}_2$, respectively. For any vector $\mathbf{z}$ in the column space of $\mathbf{W}_1$, we have $\mathbf{P}_1\mathbf{z} = 1\mathbf{z}$ and similarly for $\textbf{W}_2$, we have $\mathbf{P}_2\mathbf{z} = 1\mathbf{z}$. Therefore, for any vector $\mathbf{z}$ in the shared subspace of both matrices we have:
   \begin{equation}
       \mathbf{P}_{1,2}\mathbf{z} = \mathbf{P}_1(\mathbf{P}_2\mathbf{z}) = \mathbf{P}_1\mathbf{z} = \mathbf{z}
       \label{eq:proj}
   \end{equation}
Thus, $\mathbf{z}$ is an eigenvector of $\mathbf{P}_{1,2}$ with eigenvalue 1 and can be considered to span the shared subspace of task vector weight matrices $\textbf{W}_1$ and $\textbf{W}_2$. Meanwhile, isolated subspaces would be wiped out in the double projection $\mathbf{P}_{1,2}$ leading to a corresponding eigenvalue of 0. In other words, the eigenvalues of $\mathbf{P}_{1,2}$ are the squared cosines of the principal angles between $\mathbf{W}_1$ and $\mathbf{W}_2$. Therefore, eigenvectors corresponding to eigenvalues of 1 span aligned (shared) subspaces. We may extend this approach to more than 2 vectors by chaining the $\mathbf{P}$ matrices of all vectors considered for decomposition, i.e., $\mathbf{P}_{1,2,\cdots,k}$ for $k$ task vectors.

\subsection{Extracting Shared and Unique Components}
\label{sec:extractcomponent}
Building on Proposition \ref{theorem}, we now detail the complete procedure for extracting shared and unique components. Given the weight matrices $\{\mathbf{W}_i\}_{i=1}^k$ for $k$ task vectors, we begin by computing their singular value decompositions:
\begin{equation}
\mathbf{U}_i \Sigma_i \mathbf{V}_i^T=\mathbf{W}_i
\end{equation}
We then construct projection matrices for each task vector using the left singular vectors:
\begin{equation}
\mathbf{P}_i = \mathbf{U}_i \mathbf{U}_i^T
\end{equation}
Following \cref{eq:proj}, we chain these projections to form:
\begin{equation}
\mathbf{P}_{1,2,\dots,k} = \prod\mathbf{P}_{i=1}^{\,k}
\end{equation}
The eigendecomposition of this matrix yields eigenvalues ($\Lambda$) and eigenvectors ($\mathbf{Z}$) that characterize the shared and unique subspaces:
\begin{equation}
\mathbf{Z}\Lambda\mathbf{Z}^T = \mathbf{P}_{1,2,\dots,k}
\end{equation}
To identify the shared directions, we select eigenvectors corresponding to eigenvalues exceeding a threshold $\tau$:
\begin{equation}
\mathbf{Z}^{\text{shared}} = \mathbf{Z}[:, \mathbf{r}] \text{ where } \mathbf{r} = (\Lambda > \tau)
\end{equation}
This threshold is relaxed from the theoretical value of 1 in Proposition \ref{theorem} to account for the practical factor that task vectors from fine-tuning on similar datasets rarely align perfectly. Using these eigenvectors, we construct a projection matrix for the shared subspace:
\begin{equation}
\mathbf{P}^{\text{shared}} = \mathbf{Z}^{\text{shared}} \mathbf{Z}^{{\text{shared}}^T}
\end{equation}
The shared component of each vector is then computed as:
\begin{equation}
\mathbf{W}_i^{\text{shared}} = \mathbf{P}^{\text{shared}}\mathbf{W}_i
\end{equation}
With the unique component obtained by subtraction:
\begin{equation}
\mathbf{W}_i^{\text{unique}} = \mathbf{W}_i- \mathbf{W}_i^{\text{shared}}
\end{equation}
Since this process yields multiple shared components, one for each task vector, we combine them into a single shared task vector through a weighted average:
\begin{equation}
\mathbf{W}^{\text{shared}} = \frac{\sum_{i=1}^{k} ||\mathbf{W}_i^{\text{shared}}||_\textbf{F} \mathbf{W}_i^{\text{shared}}}{\sum^{k}_{i=1} ||\mathbf{W}_i^{\text{shared}}||_\textbf{F}}
\end{equation}
We provide results of decomposition on toy examples with known shared components between two synthetic task vectors in Appendix \ref{sec:toyexperiment} to show the recoverability of shared components in noisy conditions. We also explore an alternative decomposition method in Appendix \ref{sec:alternative}, using principal angles from QR decomposition, and show its consistency with our approach, validating its soundness.

\section{Experiments}
\label{sec:experiments}

We evaluate our method on three applications: image classification, image generation, and toxicity removal. Our method detailed in \cref{sec:extractcomponent} has a single variable, the eigenvalue threshold $\tau$, that we set to 0.85 for all experiments, which we found to work well empirically across all implementations. We present the results of our experiments for each application separately in this section alongside the related ablation studies and in-depth experiments. 
\begin{table*}
    \caption{Multi-Task merging performance with 8 datasets. Our method is applicable to task arithmetic based approaches, improving their results by providing finer control over decomposed components. *Finetuned method here represents the upper-bound for multi-task merging when only 1 dataset is targeted.}
    \label{tab:multitask}
    \centering
    \resizebox{0.9\textwidth}{!}{%
    \begin{tabular}{lccccccccc}
        \toprule
        Method & SUN397 & Cars & RESISC45 & EuroSat & SVHN & GTSRB & MNIST & DTD & Avg. \\
        \midrule
        Pre-Trained & 63.2 & 59.6 & 60.2 & 45.0 & 31.6 & 32.6 & 48.3 & 44.4 & 48.1 \\
        Fine-Tuned*  & 75.3 & 77.7 & 96.1 & 99.7 & 97.5 & 98.7 & 99.7 & 79.4 & 90.5 \\
        \midrule
        \multicolumn{10}{l}{\textit{Model Merging}} \\
        Weight Averaging & 65.3 & 63.3 & 71.4 & 73.6 & 64.2 & 52.8 & 87.5 & 50.1 & 66.0 \\
        Fisher Merging \cite{fishermerging} & \textbf{68.6} & 69.2 & 70.7 & 66.4 & 72.9 & 51.1 & 87.9 & 59.9 & 68.3 \\
        RegMean \cite{jin2023dataless} & 65.3 & 63.5 & 75.6 & 78.6 & 78.1 & 67.4 & 93.7 & 52.0 & 71.8 \\
        TIES-Merging \cite{yadav2023tiesmerging} & 65.0 & 64.3 & 74.7 & 76.8 & 81.3 & 69.4 & 96.5 & 54.3 & 72.8 \\
        \midrule
        \multicolumn{10}{l}{\textit{Task Vectors}} \\
        Task Arithmetic \cite{task_vec} & 55.3 & 54.9 & 66.7 & 77.4 & 80.2 & 69.7 & 97.3 & 50.1 & 69.0 \\
        AdaMerging (Task) \cite{Yang2023AdaMergingAM} & 58.1 & 54.5 & 67.0 & 79.2 & 79.0 & 84.6 & 93.8 & 44.6 & 70.1 \\
        AdaMerging++ (Task) \cite{Yang2023AdaMergingAM} & 60.8 & 56.9 & 73.1 & 83.4 & 87.3 & 82.4 & 95.7 & 50.1 & 73.7 \\
        AdaMerging (Layer) \cite{Yang2023AdaMergingAM} & 64.5 & 68.1 & 79.2 & 93.8 & 87.0 & 91.9 & 97.5 & 59.1 & 80.1 \\
        AdaMerging++ (Layer) \cite{Yang2023AdaMergingAM} & 66.6 & 68.3 & 82.2 & 94.2 & \textbf{89.6} & 89.0 & \textbf{98.3} & 60.6 & 81.1 \\
        \midrule
        \multicolumn{10}{l}{\textit{Our Method}} \\
        Task Arithmetic + \textbf{DeVec} & 61.9 & 61.6 & 78.1 & 82.8 & 81.5 & 80.5 & 88.5 & 53.7 & 73.6 \\
        AdaMerging (Task) + \textbf{DeVec} & 59.6 & 65.3 & 73.5 & 88.0 & 80.4 & 84.9 & 94.1 & 55.4 & 75.1 \\
        AdaMerging (Layer) + \textbf{DeVec} & 65.1 & \textbf{71.8} & \textbf{83.7} & \textbf{94.7} & 86.4 & \textbf{95.9} & 97.1 & \textbf{65.2} & \textbf{82.5} \\
        \bottomrule
    \end{tabular}}
    \vspace{-3mm}
\end{table*}
\subsection{Image Classification}
For image classification, we use OpenAI CLIP \cite{radford2021learningtransferablevisualmodels} with the ViT-B-32 architecture. We form task vectors using fine-tuned checkpoints from the following datasets: StanfordCars \cite{carsdataset} (Cars), EuroSAT \cite{helber2019eurosat}, GTSRB \cite{gtsrb}, MNIST \cite{lecun2010mnist}, RESISC45 \cite{Cheng_2017}, SVHN \cite{svhn}, CIFAR100 \cite{Krizhevsky2009LearningML}, ImageNet1k \cite{ILSVRC15} (ImageNet), Caltech256 \cite{li_andreeto_ranzato_perona_2022}, FGVCAircraft \cite{fgvcaircraft}, OxfordIIITPet (Pets) \cite{parkhi12a}, Country211 \cite{CLIP_country211}, SUN397 \cite{xiao2010}, and DTD \cite{cimpoi14describing}. A brief description of the labels of each dataset is provided in Appendix \ref{sec:datasets}. We evaluate our decomposition approach for this application in both multi-task merging (adopted from \cite{Yang2023AdaMergingAM}) and generalization scenarios to examine how vector decomposition affects model performance by providing additional degrees of freedom. These additional degrees of freedom are the shared components between each pair of available task vectors derived using our approach detailed in \cref{sec:extractcomponent}. 

\textbf{Multi-task merging}: We evaluate our approach using eight target datasets and their corresponding task vectors. Both model parameters and task vectors are frozen, with a single coefficient assigned per task vector. Only the task vector coefficients are updated during training. \cref{tab:multitask} presents results from two settings: a supervised setting, Task Arithmetic+DeVec, built on \cite{task_vec}, and an unsupervised setting, AdaMerging+DeVec, which employs entropy minimization following \cite{Yang2023AdaMergingAM}. AdaMerging distributes vector coefficients in two ways: (1) assigning a shared set of coefficients across all layers (Task-based) and (2) learning separate coefficients for each layer (Layer-based), as labeled in \cref{tab:multitask}. Additionally, AdaMerging integrates TIES-Merging with further refinements, referred to as AdaMerging++ in \cref{tab:multitask}. We compare against TIES-Merging and AdaMerging, two state-of-the-art methods designed to mitigate interference between task vectors.

Applying our method to Task Arithmetic improves accuracy across all datasets except MNIST, yielding a 4.6\% average accuracy gain. Similarly, our approach enhances Task-based AdaMerging by 5\% on average and Layer-based AdaMerging by 2.4\%. Even when AdaMerging and TIES-Merging are combined, our method further improves accuracy by 1.4\% over Layer-based AdaMerging++, demonstrating superior interference reduction through decomposition. Compared to TIES-Merging, AdaMerging (Layer) + DeVec achieves a 9.7\% accuracy boost, further validating its effectiveness. Our method also significantly outperforms RegMean, Fisher Merging, and Weight Averaging, bringing us closer to the upper bound set by full fine-tuning.

\begin{table}[h]
    \centering
    \caption{Generalization to unseen tasks EuroSAT and MNIST shows the superior performance of decomposed components.}
    \label{tab:genresults}
    \resizebox{0.47\textwidth}{!}{%
    \begin{tabular}{l c ccc}
        \toprule
        \textbf{Method} & \textbf{Seen Tasks Avg.} & \multicolumn{3}{c}{\textbf{Unseen Tasks}} \\
        \cmidrule(lr){3-5}
        & & EuroSAT & MNIST & Avg. \\
        \midrule
        Task Arithmetic~\cite{task_vec} & 70.6 & 46.2 & 77.2 & 61.7 \\
        TIES-Merging~\cite{yadav2023tiesmerging} & 69.3 & 43.3 & 75.9 & 59.6 \\
        AdaMerging~\cite{Yang2023AdaMergingAM} & 77.4 & 56.1 & 84.0 & 70.0 \\
        AdaMerging++~\cite{Yang2023AdaMergingAM} & 78.0 & 53.5 & 83.9 & 68.7 \\
        AdaMerging+DeVec & \textbf{79.0} & \textbf{60.8} & \textbf{84.7} & \textbf{72.8} \\
        \bottomrule
    \end{tabular}}
\end{table}

\textbf{Generalization}: Here we study the generalization capabilities of our approach against the SotA. We split the eight datasets into seen and unseen tasks, where MNIST and EuroSAT task vectors are removed from the pool of vectors and the remaining vectors are merged. \cref{tab:genresults} shows the average performance on the 6 seen tasks alongside the unseen and averaged unseen tasks compared against Task Arithmetic, TIES-Merging, and AdaMerging. Our method obtains significantly better results on both unseen tasks exceeding EuroSAT and MNIST performance of AdaMerging by 4.7\% and 0.7\%, respectively. Additionally, DeVec's performance of seen tasks also exceeds AdaMerging++ by 1\%. This shows how decomposing vectors into the shared and unique components allows for better generalization. 

\textbf{Ablation on Decomposed Components}: We assess the effectiveness of decomposed task vector components on downstream datasets with a relatively narrow focus. We start by adding CIFAR100, Pets, and Country211 task vectors to allow for more shared and unique components. Model performance is then evaluated in a leave-one-out setting, where each dataset is iteratively set as the target by removing its task vector. The coefficients of the remaining vectors are optimized on the target dataset’s training split, while the model and task vectors remain frozen. This experiment is repeated with different decomposed component combinations, with results shown in \cref{tab:generalization}. TV+Shared includes original task vectors with shared components as separate vectors, while Unique+Shared replaces task vectors with their unique components (\cref{sec:extractcomponent}). The top-performing approaches are TV+Shared and Shared Only, with the latter achieving the highest average accuracy—outperforming task arithmetic by 3.19\%. This suggests that shared components enable better generalization. Conversely, including unique components can degrade performance due to spurious correlations unrelated to the target dataset. Further experiments on task arithmetic generalization with shared subspaces are provided in \cref{sec:arithmetic}.

\begin{table*}[t]
    \caption{Ablation on decomposed components in a leave-on-out generalization setting. Using shared components between pairs of task vectors results in significantly better performance improvement by providing access to over or under emphasizing of shared subspaces.}
    \label{tab:generalization}
    \centering
    \resizebox{0.9\textwidth}{!}{%
    \begin{tabular}{lccccccccccc}
        \toprule
        Method & Cars & EuroSAT & GTSRB & MNIST & RESISC45 & SVHN & CIFAR100 & FGVCAircraft & Pets & Country211 & Avg. \\
        \midrule
        Zero-shot & 59.58 & 32.60 & \textbf{48.26} & 31.62 & 45.77 & 60.27 & 64.23 & 19.65 & 87.46 & 17.19 & 46.66 \\
        Task Arithmetic~\cite{task_vec} & 59.86 & 63.96 & 42.56 & 88.82 & 62.90 & 67.49 & 72.33 & 19.56 & 88.49 & 17.24 & 58.32 \\
        TV+Shared & \textbf{60.71} & \textbf{77.59} & 47.15 & 87.88 & 64.98 & 71.77 & 72.83 & 19.98 & \textbf{89.28} & \textbf{17.52} & 60.97 \\
        Unique+Shared & 60.40 & 77.07 & 46.81 & 90.75 & 64.84 & 69.72 & 71.97 & 19.65 & 89.23 & 17.49 & 60.79 \\
        Unique only & 60.12 & 62.33 & 39.34 & 81.35 & 63.60 & 54.53 & 66.20 & 20.13 & 88.03 & 17.23 & 55.29 \\
        Shared only & 60.68 & 77.56 & 47.00 & \textbf{91.70} & \textbf{65.13} & \textbf{72.78} & \textbf{73.23} & \textbf{20.25} & 89.26 & 17.46 & \textbf{61.51} \\
        \bottomrule
    \end{tabular}%
    }
    \vspace{-3mm}
\end{table*}

\subsubsection{Task arithmetic using decomposed vectors}
\label{sec:arithmetic}
We evaluate the generalizability of decomposed task vector components in two settings: (1) large-scale image classification (ImageNet, SUN397, CIFAR100, Caltech256) to assess shared and unique component effects when tasks are broadly similar, and (2) small-scale digit/letter classification (MNIST, GTSRB, SVHN) where tasks are more specific. For large datasets, we extract the shared component from SUN397, CIFAR100, and Caltech256 using three projection matrices instead of pairwise decomposition (\cref{sec:extractcomponent}). Unique components are obtained by subtracting the shared component. Task arithmetic is then performed by adding these vectors to a pretrained model and evaluating ImageNet performance—without an ImageNet task vector.

\Cref{fig:imagenet_arithmetic} shows that unique components have little impact, while the shared component improves accuracy by over 0.5\%. Of the original vectors, only Caltech256 contributes positively, suggesting large dataset task vectors do not necessarily generalize well. However, extracting shared subspaces captures transferable information. A similar experiment on a noisy adversarial dataset (Appendix \ref{sec:noisydataset}) further supports the robustness of shared components.
\begin{figure}[t]
    \vspace{-2mm}
    \centering
    \includegraphics[width=0.45\textwidth]{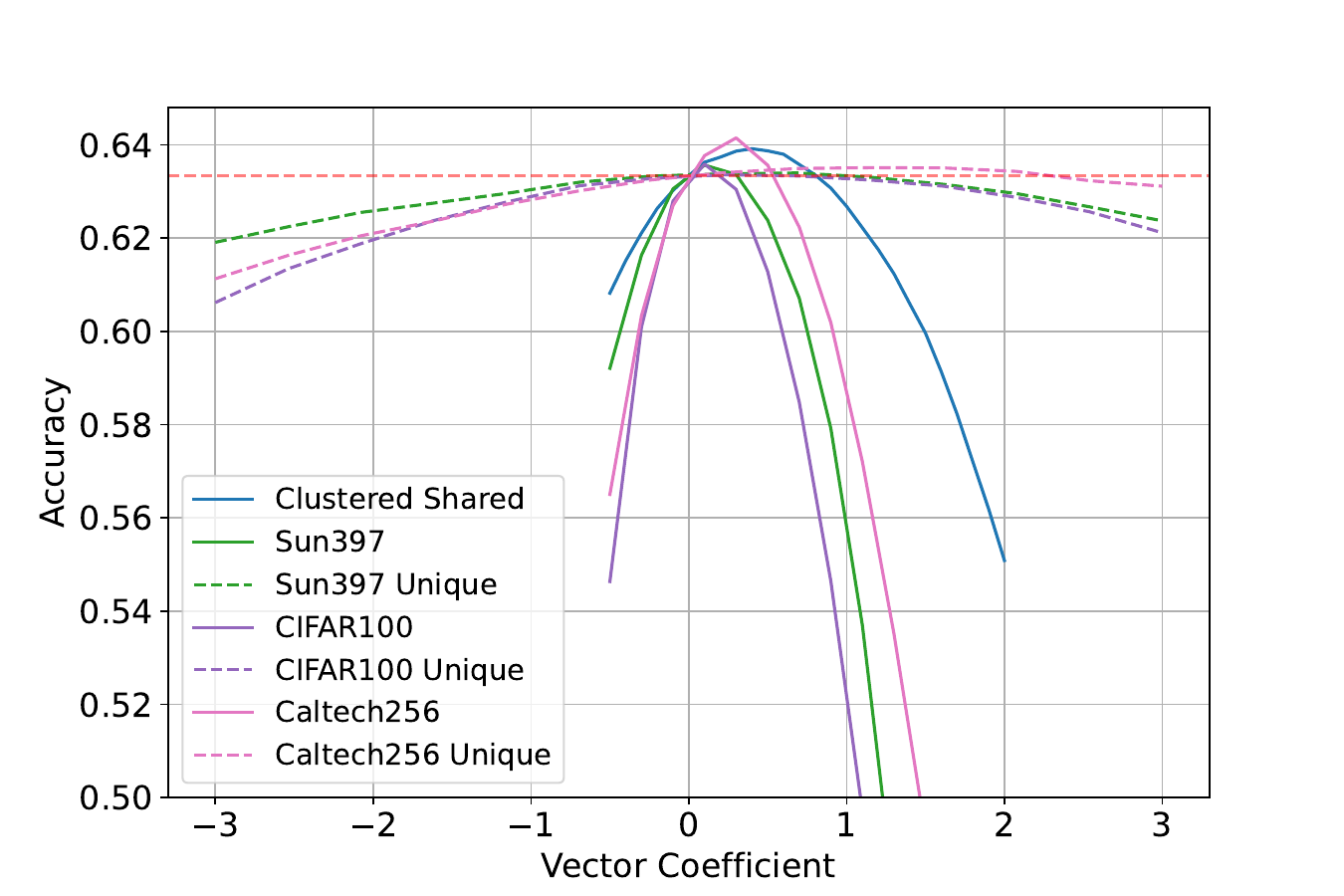}
    \vspace{-1mm}
    \caption{Arithmetic on ImageNet shows the shared component between task vectors from other large datasets consists of generalizable information allowing for accuracy improvement. However, unique components have negligible effects since the information embedded in unique vectors is related to the vector's own dataset.}
    \label{fig:imagenet_arithmetic}
    \vspace{-2mm}
\end{figure}
Repeating this with GTSRB as the target and MNIST/SVHN as task vectors, the shared component improves accuracy beyond either vector alone. After decomposition, the unique components have smaller magnitudes than the shared one, indicating substantial overlap in subspaces. In contrast, large dataset task vectors retain more dataset-specific information despite having more classes, suggesting they do not necessarily form similar subspaces during fine-tuning.
\begin{figure}[t]
    \vspace{-2mm}
    \centering
    \includegraphics[width=0.45\textwidth]{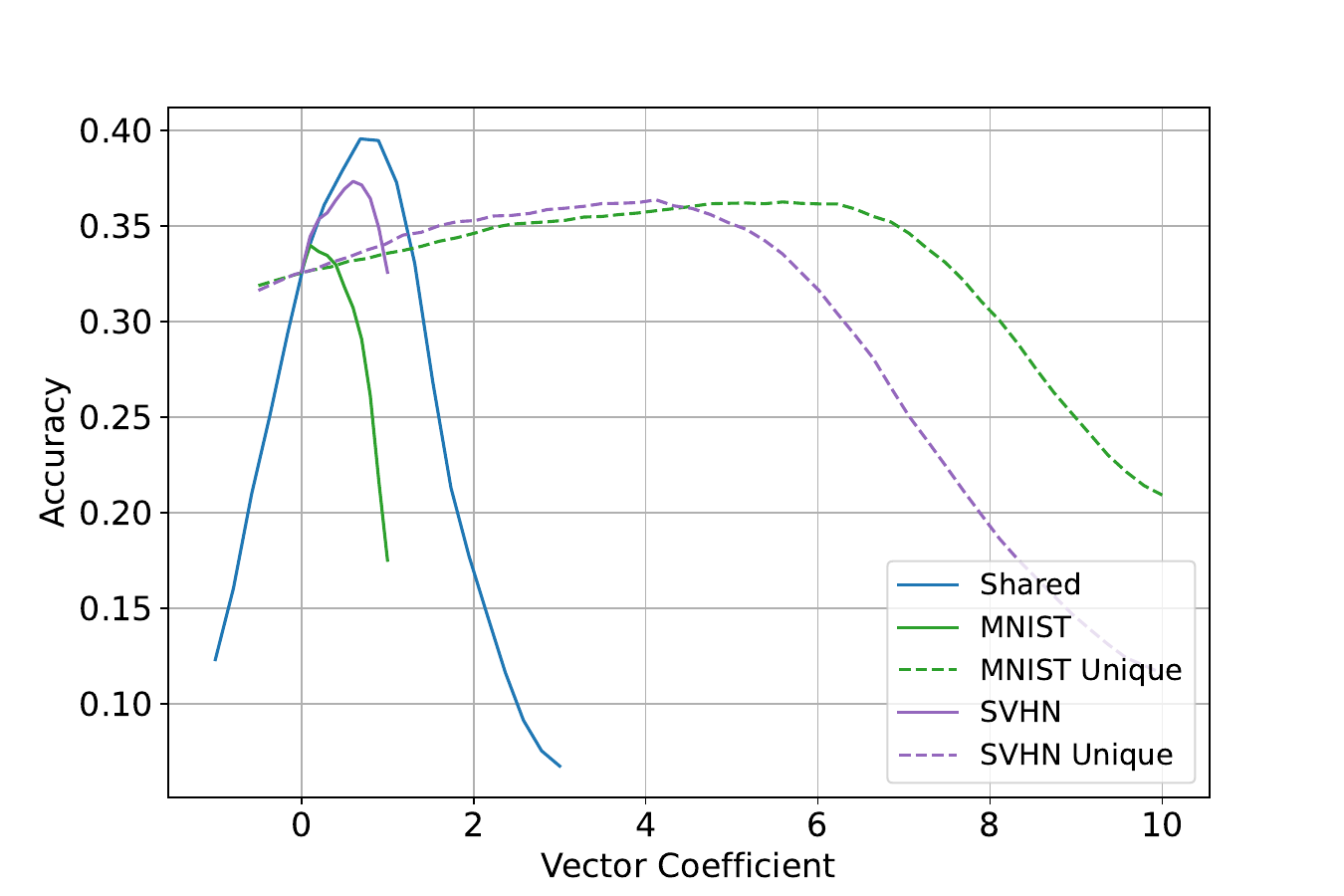}
    \vspace{-1mm}
    \caption{Arithmetic on GTSRB shows how the shared component between MNIST and SVHN improves accuracy and unique components fail to have meaningful effects.}
    \label{fig:imagenet_arithmetic}
    \vspace{-3mm}
\end{figure}
\subsection{Image Generation}
\label{sec:imagegeneration}
\begin{figure}[bht]
    \centering
    \begin{subfigure}[b]{0.236\textwidth}
        \centering
        \includegraphics[width=\textwidth]{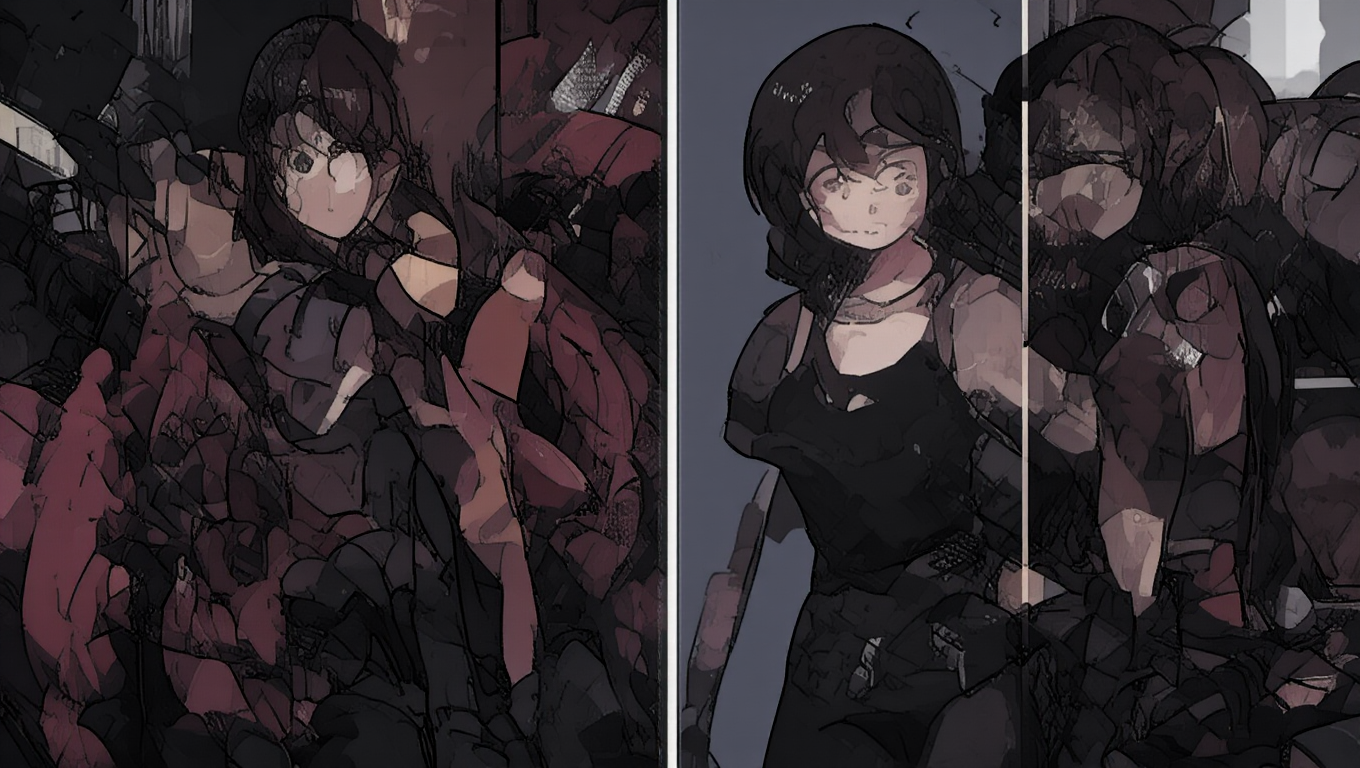}
        \label{fig:fig1}
    \end{subfigure}
    \hfill
    \begin{subfigure}[b]{0.236\textwidth}
        \centering
        \includegraphics[width=\textwidth]{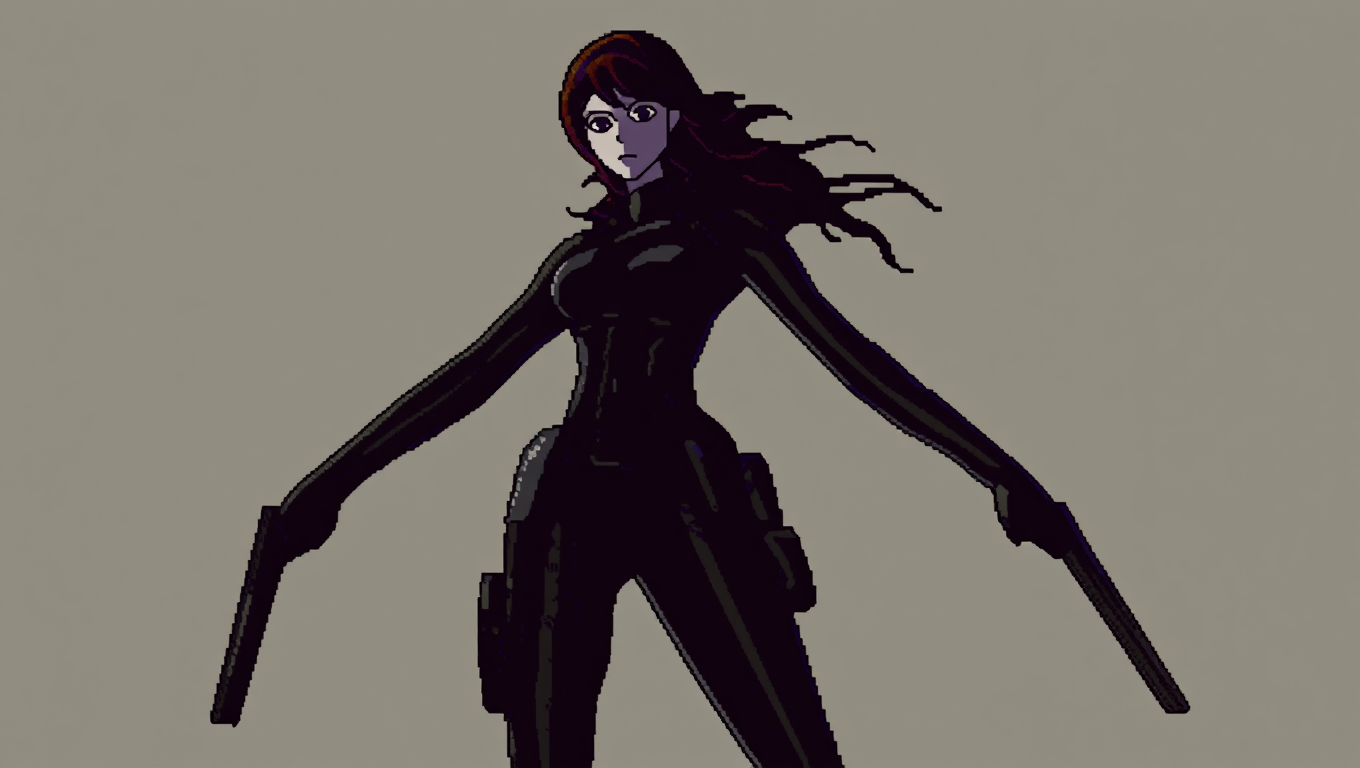}
        \label{fig:fig2}
    \end{subfigure}
    
    \vspace{-1em}
    
    \begin{subfigure}[b]{0.236\textwidth}
        \centering
        \includegraphics[width=\textwidth]{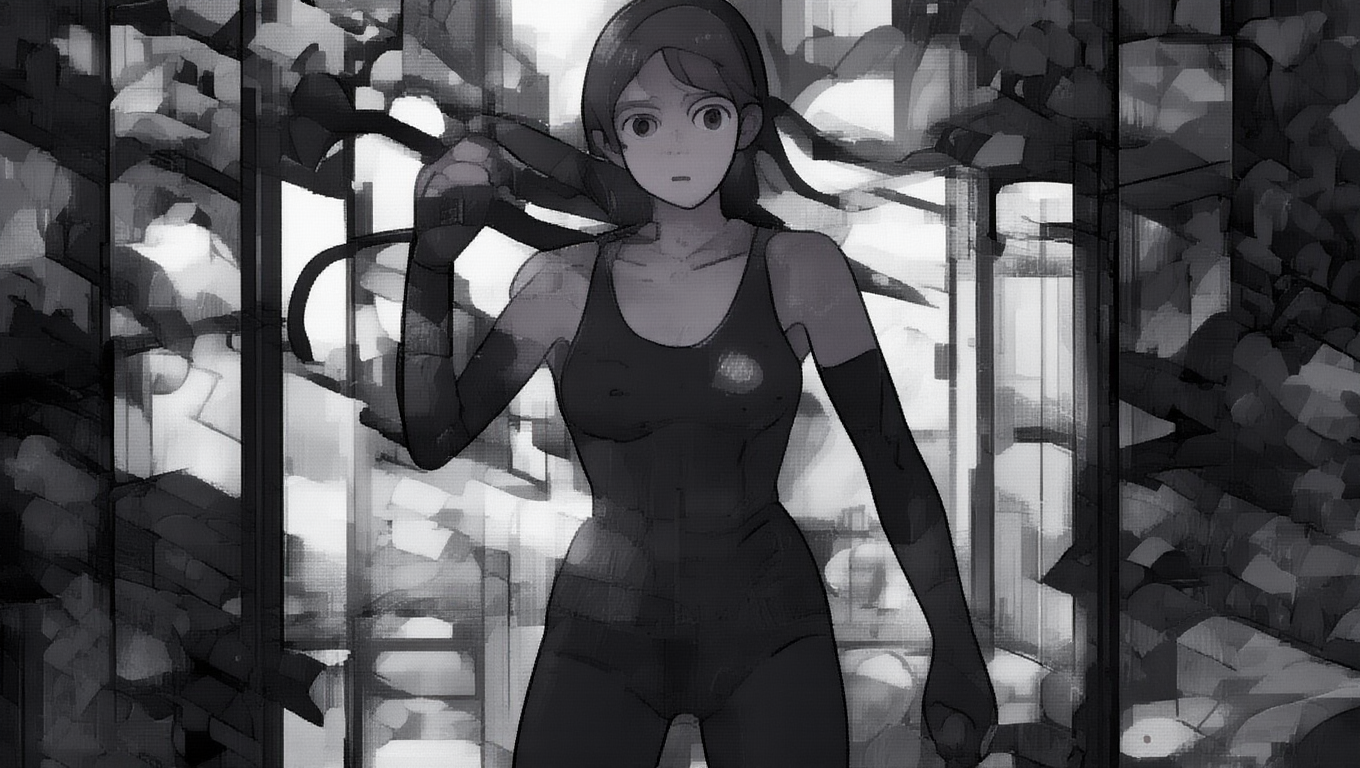}
        \label{fig:fig3}
    \end{subfigure}
    \hfill
    \begin{subfigure}[b]{0.236\textwidth}
        \centering
        \includegraphics[width=\textwidth]{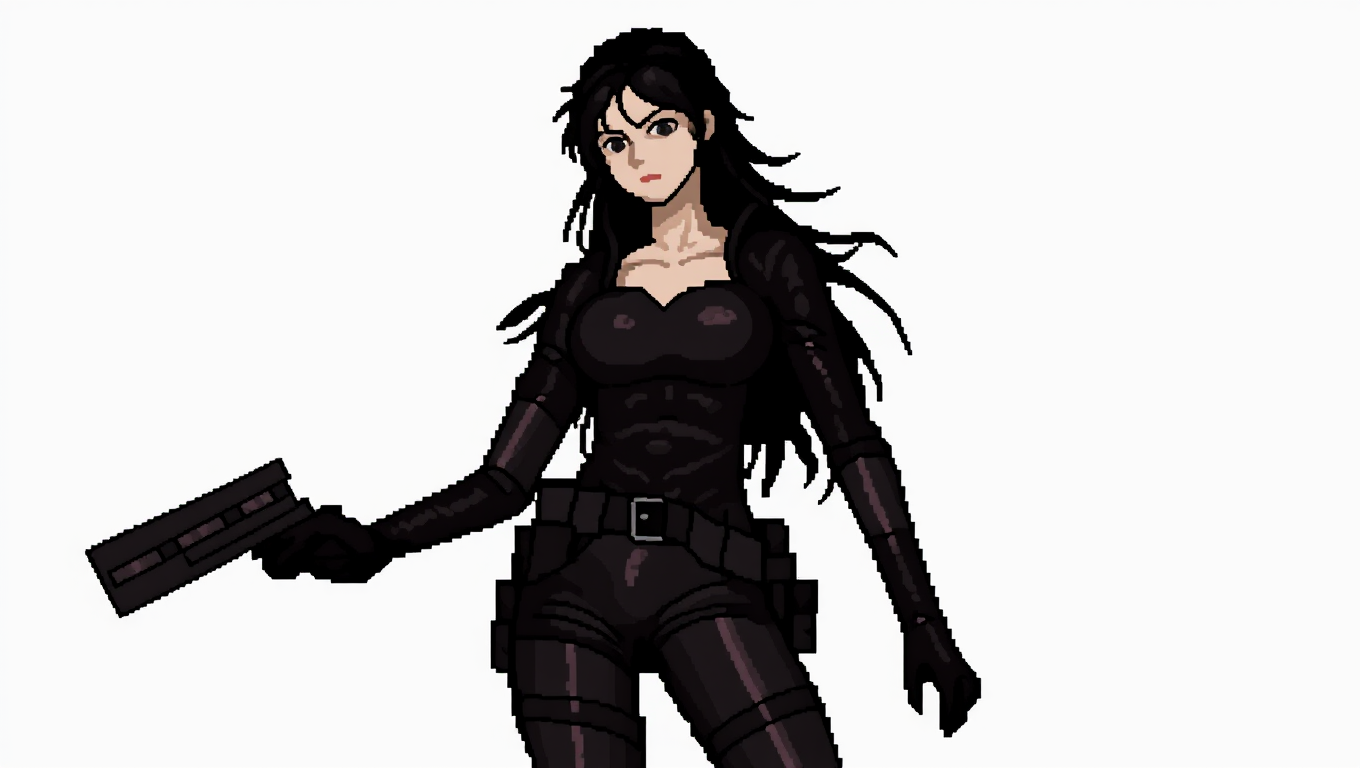}
        \label{fig:fig4}
    \end{subfigure}
    
    \vspace{-1em}
    
    \begin{subfigure}[b]{0.236\textwidth}
        \centering
        \includegraphics[width=\textwidth]{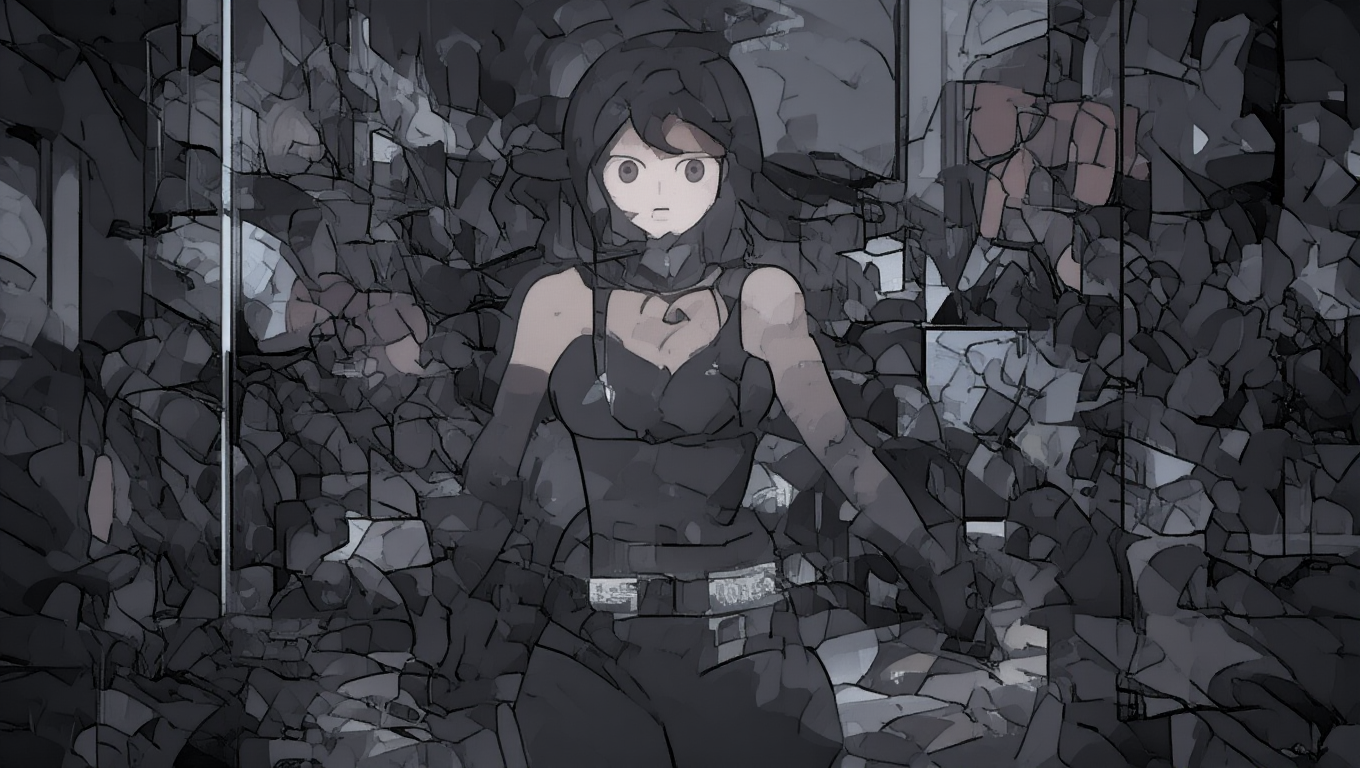}
        \caption{Vanilla task vectors}
        \label{fig:fig5}
    \end{subfigure}
    \hfill
    \begin{subfigure}[b]{0.236\textwidth}
        \centering
        \includegraphics[width=\textwidth]{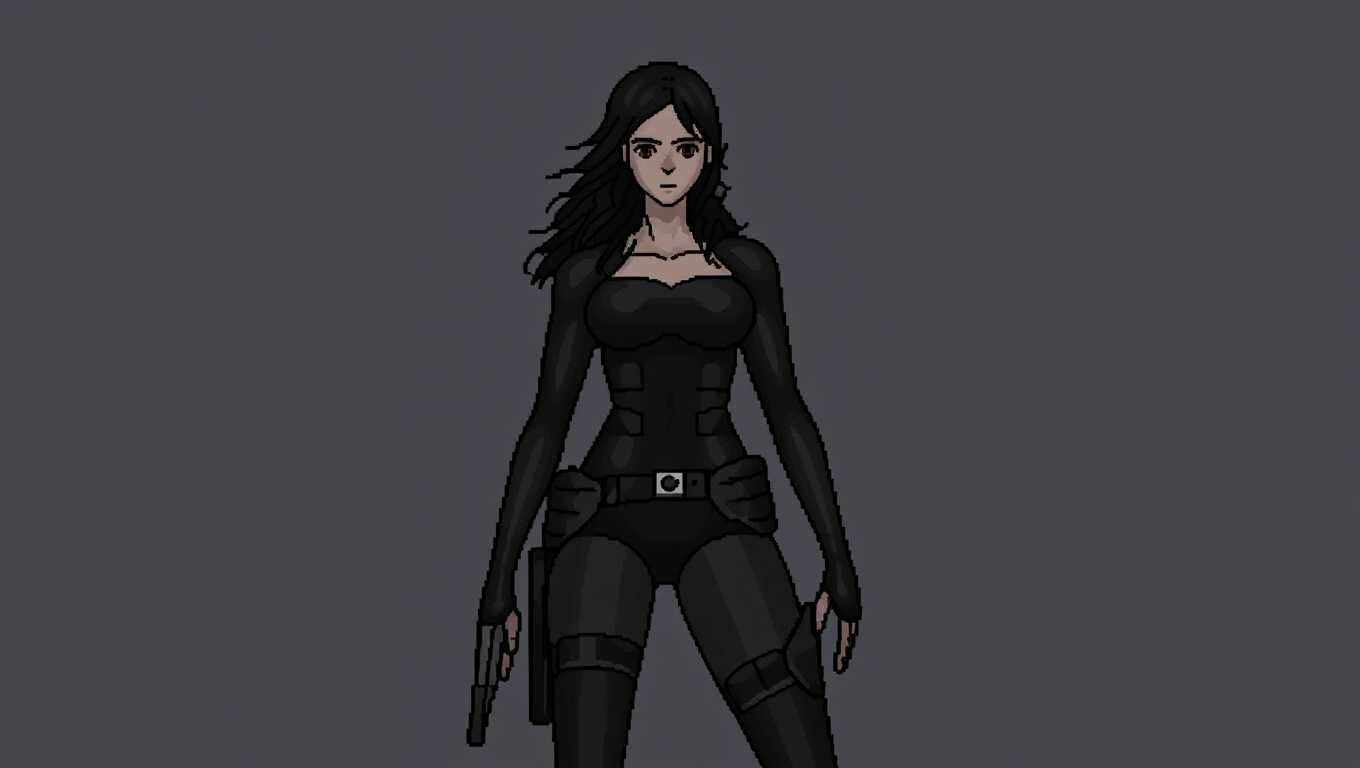}
        \caption{Unique vectors (ours)}
        \label{fig:fig6}
    \end{subfigure}
    
    \caption{Text-to-image model outputs with prompt "pixelart of athletic adult woman dressed in a sleek black and dark grey jumpsuit, with a utility belt around her waist, armed with an arsenal of vampire-hunting gadgets".}
    \label{fig:texttoimage}
    \vspace{-3.5mm}
\end{figure}

For image generation, we explore style mixing using the Flux Schnell 1.0 text-to-image model \cite{FLUX1_schnell}. Since fine-tuned models for specific styles are trained on limited data (often $<100$ images), shared components between vectors contain little meaningful information for generation quality. Mixing multiple style vectors can degrade outputs due to conflicts and overemphasis on shared components, while unique components retain style-specific features. To address this, we extract and combine only unique components. Here, style vector coefficients are not learned but set to 1 for mixing. \Cref{fig:texttoimage} visualizes the results of mixing four style vectors—three trained on different animation styles and one on pixel art. Individual style generations are shown in Appendix \ref{sec:eachstyle}. We chose multiple animation styles to test whether their shared component dominates, while the pixel art vector helps assess if it is overruled by the others. \Cref{fig:fig5} shows that mixing original vectors results in degraded outputs due to overemphasis on shared components, while using only unique components allows for proper style mixing. Additionally, mixing original vectors eliminates pixelation, showing that the pixel art vector is undermined. In contrast, mixing unique components preserves all styles. Additional results with different prompts and style combinations are provided in Appendix \ref{sec:morestylemixing}.

\subsection{Toxicity Removal}
\label{sec:toxicityremoval}
We evaluate our method’s ability to reduce toxic text generation in LLAMA-2-7B \cite{touvron2023llama2openfoundation}. First, we fine-tune the model on ToxiGen \cite{toxigen} and RealToxicityPrompts \cite{realtoxicityprompts} (details in Appendix \ref{sec:llamafinetune}) to obtain the corresponding toxic vectors. Using our approach iteratively with seven additional fine-tuned checkpoints from HuggingFace (listed in Appendix \ref{sec:llamacheckpoints}), we extract the unique toxic component by removing shared components between the toxic vector and other task vectors. These retrieved vectors, trained on physics books, Roman novels, and instruction datasets, are unrelated to toxic text generation, ensuring that only toxicity-related information is isolated. We then subtract this component from the pre-trained model and assess both toxicity reduction on benchmark test sets and performance on control tasks (reasoning and general knowledge), comparing against state-of-the-art model editing methods. Therefore, the extracted toxic component's coefficient is negative.

\begin{table*}[t]
    \centering
    \caption{\small Comparison to SotA toxicity removal methods. Lower scores ($\downarrow$) are better for ToxiGen and RealToxicity (RealTox in table), indicating reduced toxic generation. Higher scores ($\uparrow$) are better for control tasks, showing minimal impact from toxicity removal. \underline{Underlined} values indicate significant performance drops. Accuracy is reported for each control task per toxic task vector. *Baselines from \cite{wang2024modelsurgerymodulatingllms}.}
    \label{tab:toxicity}
    \resizebox{0.84\textwidth}{!}{%
    \begin{tabular}{c|cc|ccccc}
    \toprule
    Methods & \multicolumn{2}{c|}{\textbf{Toxicity}} & \multicolumn{5}{c}{\textbf{Control Tasks}} \\
    \cmidrule(l){2-3} \cmidrule(l){4-8}
    & ToxiGen ($\downarrow$) & RealTox ($\downarrow$) & GSM8K ($\uparrow$) & BBH ($\uparrow$) & MMLU ($\uparrow$) & TydiQA ($\uparrow$) & Avg. ($\uparrow$) \\
    \midrule
    LLaMA2-7B & 79.1 & 51.4 & 14.6 & 39.0 & 41.7 & 48.1 & 35.9 \\
    \midrule
    SFT & 86.7 & 34.4 & $\underline{8.95}$ & \underline{27.5} & \underline{32.3} & \underline{22.8} & \underline{22.9} \\
    Task Vector~\cite{task_vec} & 73.1 & 17.3 & 14.7 & \underline{30.1} & 37.8 & \underline{43.4} & \underline{31.5} \\
    Contrastive Decoding* & 73.5 & 14.6 & 13.0 & 39.0 & 41.2 & 49.1 & 35.6 \\
    Safe Activation* & 71.9 & 38.9 & \underline{10.3} & 38.5 & 40.9 & 46.9 & 34.2 \\
    Feature Subtraction* & 53.5 & 15.9 & 15.5 & \underline{15.7} & \underline{33.7} & \underline{21.3} & \underline{21.6} \\
    Model Surgery~\cite{wang2024modelsurgerymodulatingllms} & 39.9 & \textbf{5.17} & 14.4 & 37.7 & 41.7 & 45.6 & 34.9 \\
    \midrule
    \textbf{DeVec (Ours)} & \textbf{32.1} & 5.90 & 14.9/14.5 & 37.8/36.3 & 39.5/39.7 & 44.4/48.1 & 34.2/34.7 \\
    \bottomrule
    \end{tabular}%
    }
    \vspace{-1mm}
\end{table*}

\Cref{tab:toxicity} presents our results. We bring the results on the SotA and baselines from \cite{wang2024modelsurgerymodulatingllms}. Our approach outperforms Model Surgery, the state-of-the-art method, on ToxiGen and achieves competitive results on RealToxicityPrompts. Unlike \cite{wang2024modelsurgerymodulatingllms}, which requires linear probing of intermediate features, our method achieves these results using only projection and linear operations, without accessing hidden states. Notably, vanilla task vectors fail to reduce toxicity effectively due to their shared component, which prevents toxicity removal without harming performance on control tasks. In contrast, our approach preserves accuracy on control tasks while effectively mitigating toxicity, demonstrating its robustness in model editing.

\subsubsection{Negation Coefficient}
\begin{figure}[t]
    \centering
    \begin{subfigure}[b]{0.5\textwidth}
        \centering
        \includegraphics[width=0.9\textwidth]{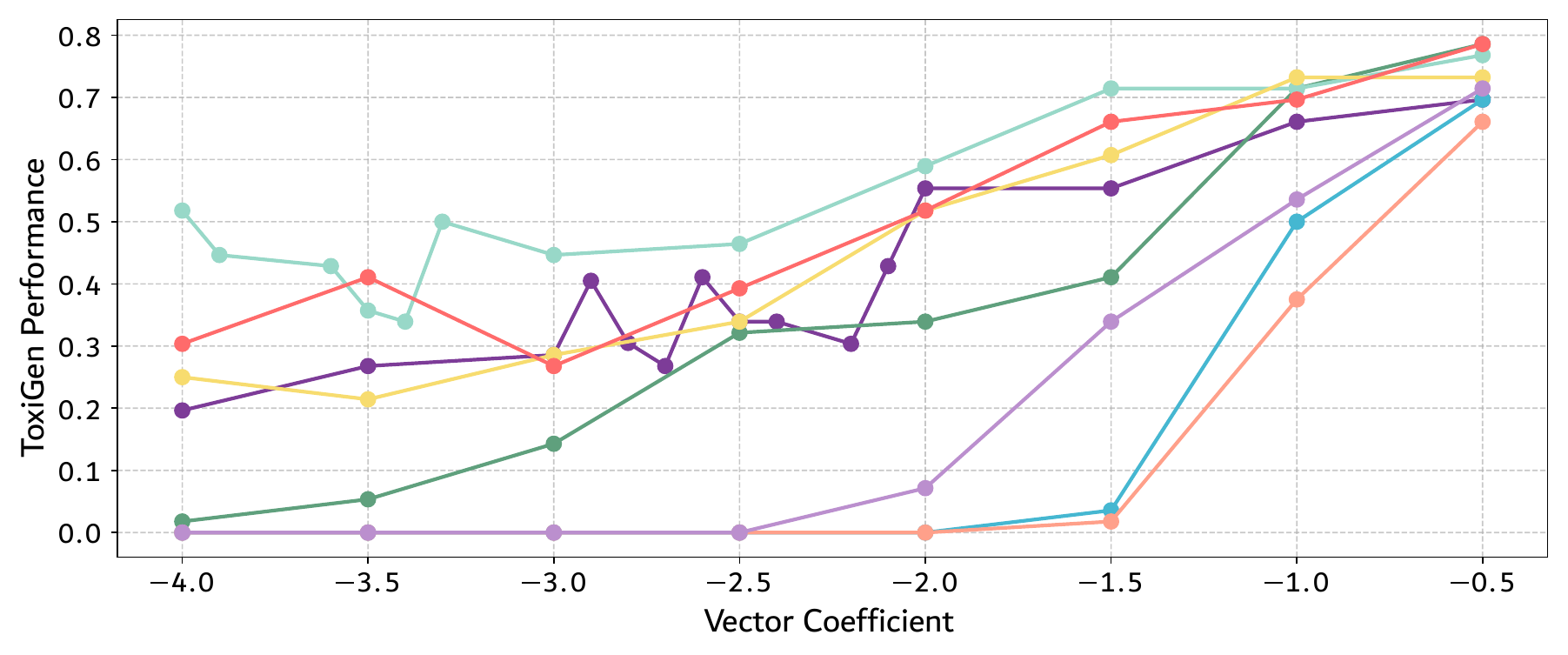}
        \caption{Toxicity over vector coefficient}
        \label{fig:negation_figure1}
    \end{subfigure}
    \vspace{1em}
    \begin{subfigure}[b]{0.5\textwidth}
        \centering
        \includegraphics[width=0.9\textwidth]{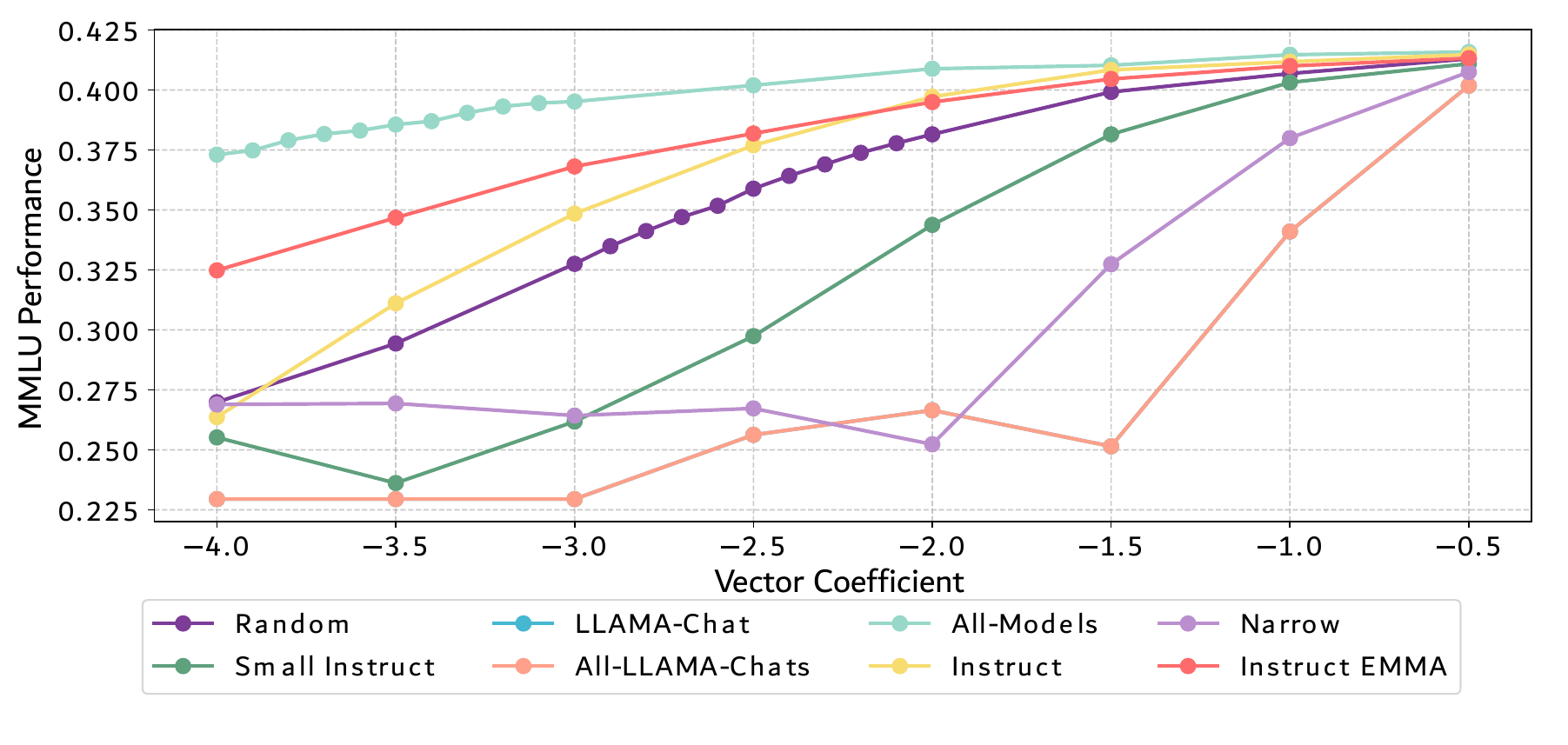}
        \caption{MMLU performance over vector coefficient}
        \label{fig:negation_figure2}
    \end{subfigure}
    \vspace{-8mm}
    \caption{Generation toxicity is reduced with a larger negation coefficient but past a threshold, performance on control tasks degrades. The unique vector resulting from a larger pool of task vectors strikes a balance between toxicity and performance. See Appendix \ref{sec:llamacheckpoints} for the naming of task vector pools.}
    \label{fig:negation}
    \vspace{-6mm}
\end{figure}

We conduct an ablation study on the negation coefficient in the toxicity removal experiment to analyze how negating the toxic component affects both toxicity reduction and control task performance. \cref{fig:negation} visualizes these changes. Each line represents negation using the unique component obtained from a \emph{subset} of available task vectors (unlike \Cref{tab:toxicity}, which used all retrieved vectors). The best balance between toxicity reduction and control task stability is achieved when using all task vectors (named `All-Models`). In contrast, the worst performance on control tasks occurs when only two vectors are used to extract the unique toxic component (named `All-LLAMA-Chat`). As the number of task vectors used for obtaining the unique toxic component increases, the decline in control task accuracy slows. This shows that removing shared components between the toxic and other vectors isolates embedded knowledge, refining the unique vector to better capture the toxic information.

\subsubsection{Decomposition Coefficient}
\begin{figure}[t]
    \centering
    \begin{subfigure}[b]{0.5\textwidth}
        \centering
        \includegraphics[width=0.9\textwidth]{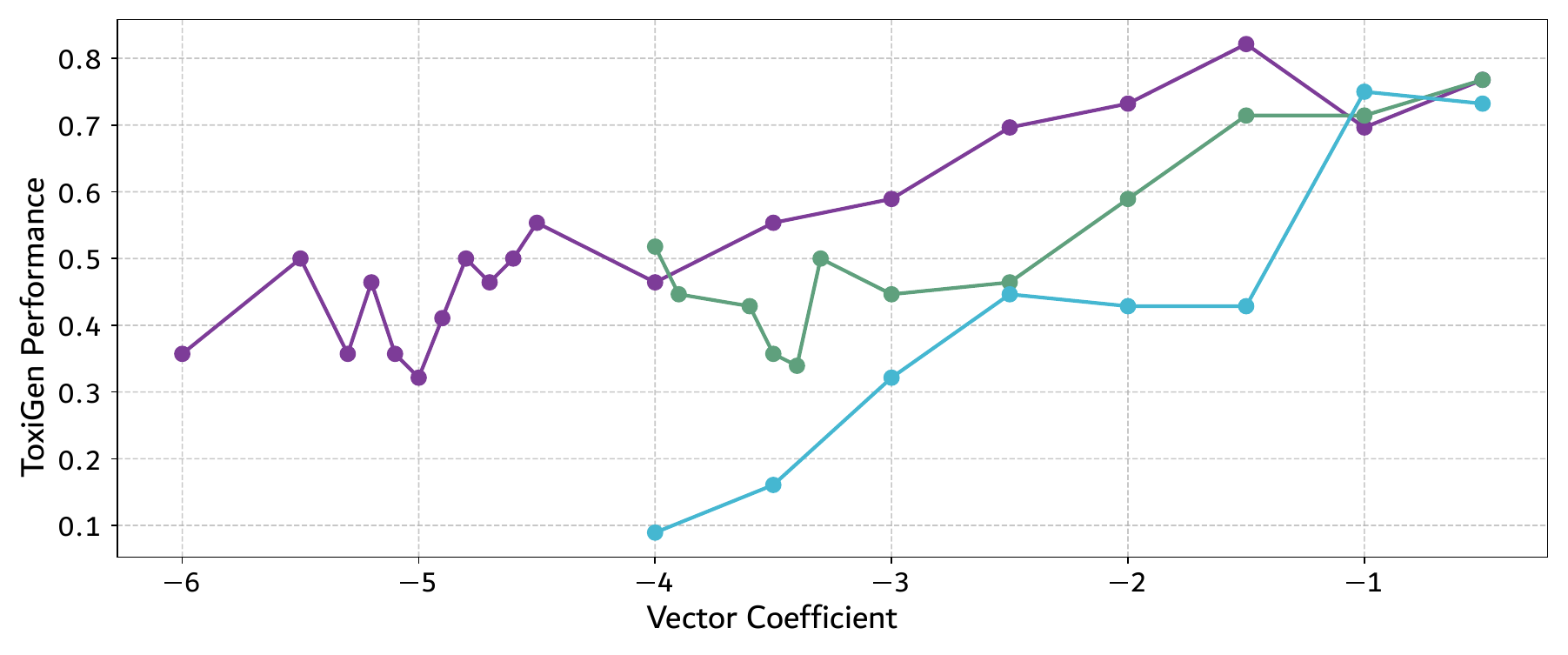}
        \caption{Toxicity over eigenvalue thresholds}
        \label{fig:decompcoeff_figure1}
    \end{subfigure}
    \vspace{1em}
    \begin{subfigure}[b]{0.5\textwidth}
        \centering
        \includegraphics[width=0.9\textwidth]{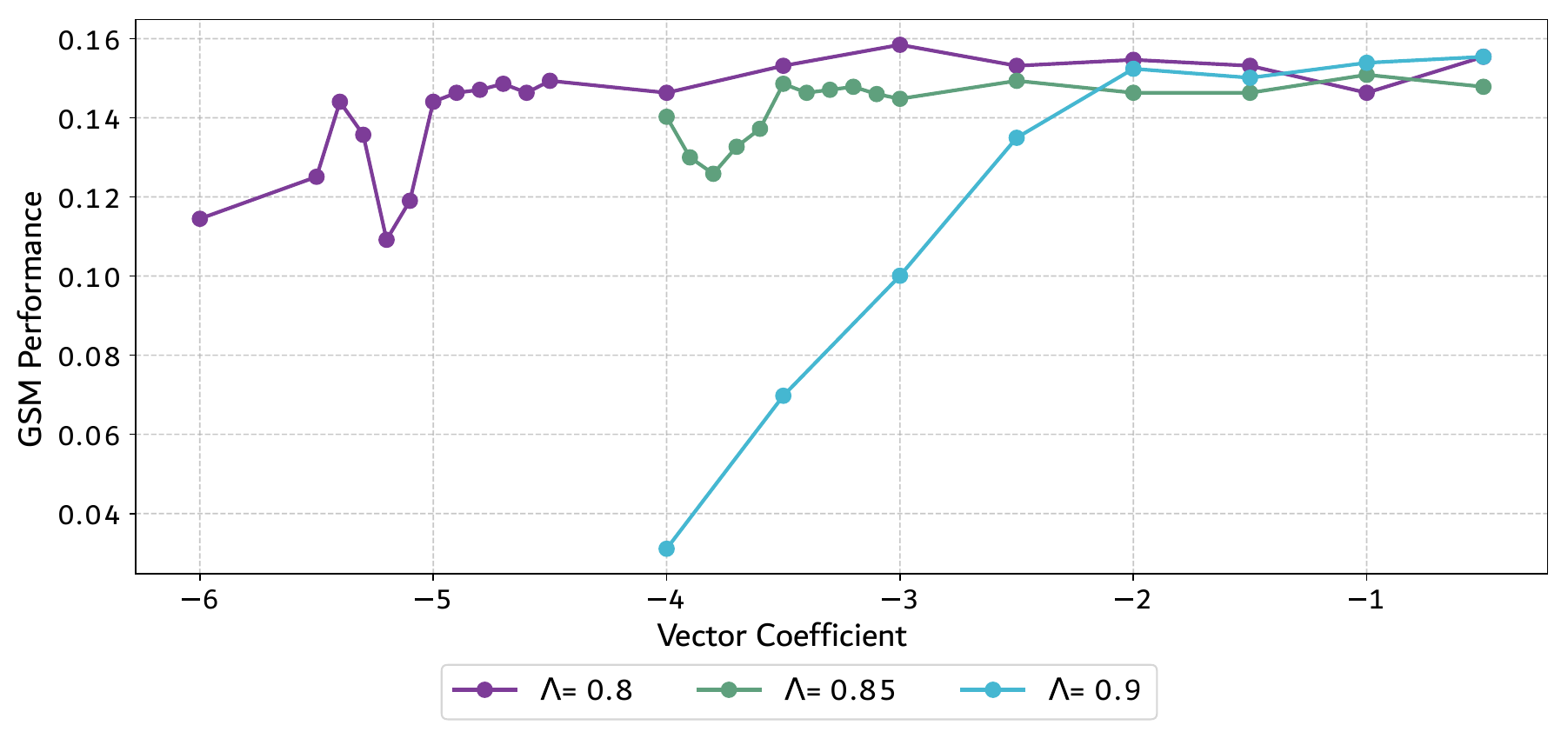}
        \caption{GSM performance over eigenvalue thresholds}
        \label{fig:decompcoeff_figure2}
    \end{subfigure}
    \vspace{-8mm}
    \caption{Too high of an eigenvalue threshold leads to shared subspaces being considered unique since vectors from different datasets will not have perfectly aligned shared subspaces.}
    \label{fig:decompcoeff}
    \vspace{-6mm}
\end{figure}

Here we study the variations in eigenvalue threshold $\Lambda$. For the main experiments of every application, so far, we have used a threshold of 0.85. \cref{fig:decompcoeff} shows that by choosing a threshold value too high, the shared components may be mistaken for unique ones. In such a case, the accuracy on toxic vectors does drop to some extent (although not as much as the other threshold values), but the accuracy on control tasks falls sharply with a stronger negation coefficient. This is while a threshold value of 0.8 leads to unique subspaces counting as shared ones where the toxicity remains relatively higher (in the range of -4.5 to -5) during a stable control task performance. Therefore, a threshold of 0.85 seems to be a suitable choice balancing the scales between unique and shared components.
\section{Conclusion}
\label{sec:conclusion}

We introduced a novel method to decompose task vectors—the difference between fine-tuned and pre-trained model parameters-into components that are either shared across multiple tasks or unique to specific behaviors. our approach isolates these subspaces, enabling more effective arithmetic operations without interference from overlapping concepts. Our method was validated across three domains. In image classification, isolating shared components improved multi-task merging and generalization, outperforming state-of-the-art methods by 5\% on average. For image generation, removing shared components enabled clean style mixing, avoiding the degradation seen with standard style vector combinations. In language models, our approach achieved state-of-the-art toxicity removal via unique component negation, preserving performance on control tasks and significantly outperforming previous methods. Our findings highlight key insights: shared components often enhance transfer learning, unique component isolation prevents concept interference, and decomposition effectiveness scales with the task vector pool size. These insights pave the way for advanced model editing, deeper analysis of task relationships, and a better understanding of transfer learning dynamics in neural networks.
{
    \small
    \bibliographystyle{ieeenat_fullname}
    \bibliography{main}
\clearpage
\setcounter{page}{1}
\maketitlesupplementary

\section{Alternative method for task vector decomposition}
\label{sec:alternative}
Our proposed approach uses projection matrices onto the column space of each task vector to identify subspaces from each task vector that are aligned past a threshold with the subspaces of another task vector. An alternative method is brought here from \cite{golub_matrix_2013} that relies on principal angles between the weight matrices of the task vectors:
\begin{theorem}[Subspace Intersection \mbox{\cite{golub_matrix_2013}}] 
Given two matrices $A\in\mathbb{R}^{n \times p}$ and $B\in\mathbb{R}^{n \times q}$, spanning $F$ and $G$ in $\mathbb{R}^m$ whose dimensions satisfy
\begin{equation}
p=\operatorname{dim}(F) \geq \operatorname{dim}(G)=q \geq 1
\end{equation}
The principal angles $\theta\in \{\theta_1, \, \theta_2, \cdots, \theta_q\}$ (ordered  by $0 \leq \theta_1 \leq \cdots \leq \theta_q \leq \pi / 2$) alongside the corresponding principal vectors $f_i$ and $g_i$ are defined recursively as \cite{golub_matrix_2013}
\begin{equation}
\cos \left(\theta_k\right)=f_k^T g_k=\max _{\substack{f \in F,\|f\|_2=1 \\ f^T\left[f_1, \ldots, f_{k-1}\right]=0}} \max _{\substack{g \in G,\|g\|_2=1 \\ g^T\left[g_1, \ldots, g_{k-1}\right]=0}} f^T g .
\label{eq:principalangles}
\end{equation}
\label{th:intersection}
\end{theorem}
For the proof, we refer to \cite{golub_matrix_2013}. By thresholding the values from \ref{eq:principalangles}, we may label subspaces as shared following the same process as \cref{sec:extractcomponent}. Our reason for not using this approach is two-fold. Projection matrices allow for the expansion of our work into activation-based task arithmetic \cite{hendel-etal-2023-context} where the projection matrices of the feature vectors have been used in recent work to show that task arithmetic is not restricted to the parameter space. Therefore, the projection matrices from our work can easily be combined with projection matrices of data to make task arithmetic operate fluently in both parameter and activation space. This approach cannot be adopted when using subspace intersection used in Theorem \ref{th:intersection}. Additionally, our approach relies on Singular Value Decomposition (SVD) to calculate the projection matrices instead of QR decomposition for Theorem \ref{th:intersection}. Efficient alternatives of SVD are more widely available and directly implemented in most commonly used libraries.

Nevertheless, we provide a comparison of the two approaches. In Figure \ref{fig:qrcomparison}, we visualize a histogram of the principal angles (in radians) between shared subspaces extracted for two generated task vectors from both approaches over 100 runs. As evident in this figure, the subspaces deemed shared for both approaches are less than 6 degrees apart.

\begin{figure}[h]
    \centering
    \includegraphics[width=0.45\textwidth]{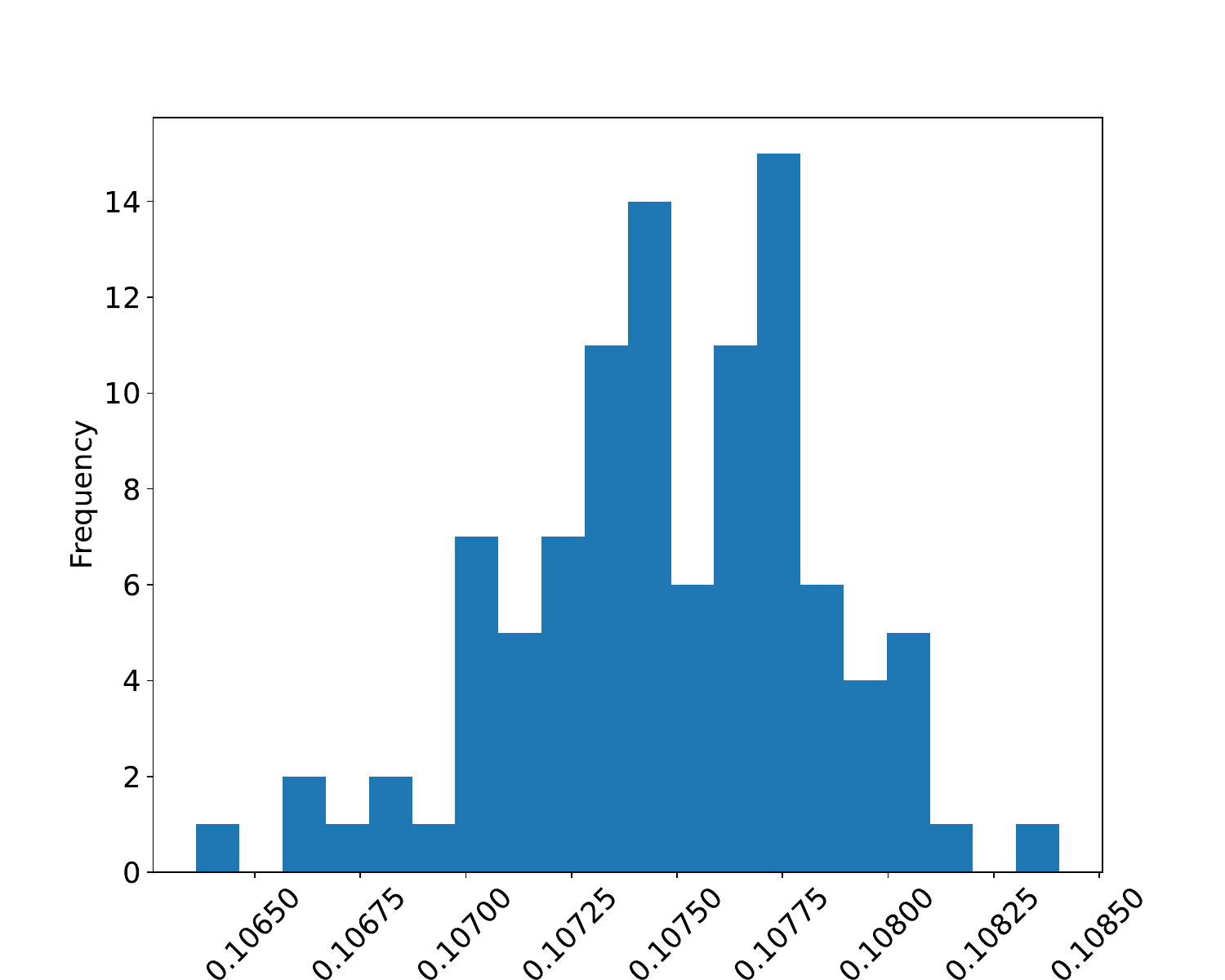}
    \caption{Principal angles between the shared components extracted by Theorem \ref{th:intersection} and Proposition \ref{theorem}. Extracted components are close to each other verifying our proposed approach.}
    \label{fig:qrcomparison}
\end{figure}

\section{Toy Experiment}
\label{sec:toyexperiment}
We validate our approach through visualization of eigenvalues and principal angles of shared components in simulated experiments using generated matrices with and without the presence of noise. Figure \ref{fig:eigenvalues} shows the eigenvalues of the $P_{1,2}$ matrix in a toy experiment. We generate two matrices with 100 known shared subspaces. Without the presence of any noise, all the shared subspaces can be recovered as shown in Figure \ref{fig:eigenvalues}. Figure \ref{fig:noisyangles} visualizes the mean of the principal angles between the true and recovered shared subspaces across different levels of noise. This value stays below 30 degrees with considerable noise levels.

\begin{figure}[h]
    \centering
    \begin{subfigure}[b]{0.5\textwidth}
        \centering
        \includegraphics[width=0.99\textwidth]{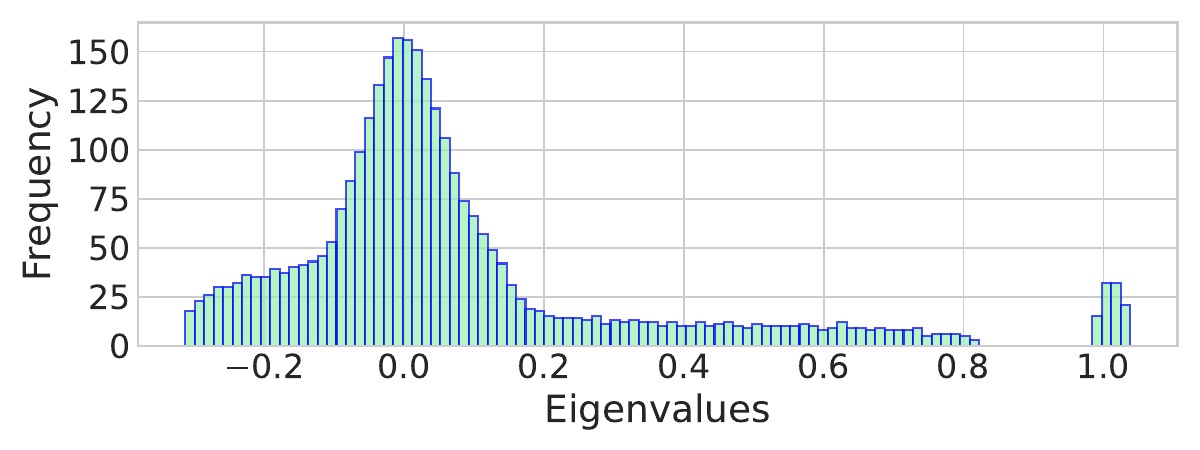}
        \caption{Histogram of Eigenvalues}
        \label{fig:eigenvalues}
    \end{subfigure}
    \vspace{1em}
    \begin{subfigure}[b]{0.5\textwidth}
        \centering
        \includegraphics[width=0.99\textwidth]{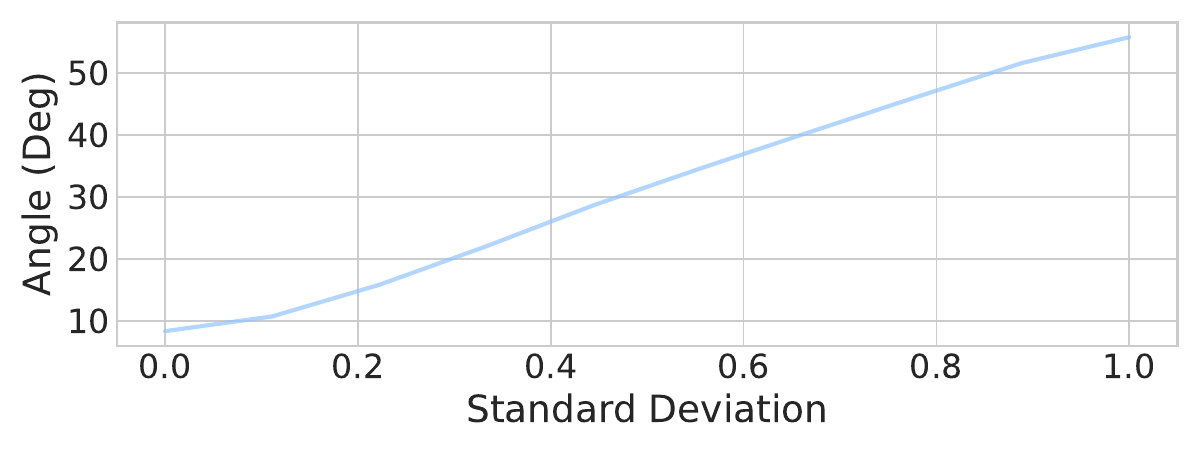}
        \caption{Mean principal angles across noise levels}
        \label{fig:noisyangles}
    \end{subfigure}
    \caption{\textbf{(a)} The eigenvalues of $P_{AB}$ are comprised of two clusters with the eigenvalues of shared subspaces clustering around 1. \textbf{(b)} Mean of principal angles between the true shared subspaces and recovered subspaces across different noise levels showing the potential of recovering the true shared subspaces to an appropriate extent with less than 30 degrees of error below Gaussian noise with $\sigma=0.4$.}
    \label{fig:both_figures}
\end{figure}

\section{Description of used datasets}
\label{sec:datasets}
Here we provide a brief description of the labels in each dataset used in our experiments. For vision datasets, we use the following:
\begin{itemize}
    \item StanfordCars (Cars) \cite{carsdataset}: Images of 196 cars from different manufacturers made in years ranging from 1991 to 2012.
    \item EuroSAT \cite{helber2019eurosat}: Labeled satellite images of Earth captured by the Sentinel-2 satellite, spanning 10 different land use and land cover classes like residential areas, forests, and industrial zones.
    \item GTSRB \cite{gtsrb}: The German Traffic Sign Recognition Benchmark (GTSRB) is a dataset of real-world traffic sign images across 43 different sign classes, captured under varying lighting and weather conditions, with images varying in size and coming from German roads.
    \item MNIST \cite{lecun2010mnist}: Grayscale images of handwritten digits.
    \item RESISC45 \cite{Cheng_2017}: High-resolution remote sensing images across 45 scene categories (like airport, forest, harbor).
    \item SVHN \cite{svhn}: Street View House Numbers (SVHN) dataset is made of real-world digit images extracted from Google Street View photos of house numbers, with each digit being cropped and labeled.
    \item CIFAR100 \cite{Krizhevsky2009LearningML}: Small color images divided into 100 fine-grained classes (grouped into 20 superclasses), with 600 images per class at 32x32 pixels, designed to test hierarchical classification and fine-grained recognition algorithms.
    \item ImageNet1k (ImageNet) \cite{ILSVRC15}: Large-scale dataset containing approximately 1.2 million training images across 1,000 object categories, serving as a foundational benchmark for computer vision models since its introduction.
    \item  Caltech101 \cite{li_andreeto_ranzato_perona_2022}: The Caltech-101 dataset contains 101 distinct object categories plus a background class, with about 40 to 800 images per category, making it one of the earlier benchmark datasets for object recognition despite its relatively small size by modern standards.
    \item  FGVC-Aircraft \cite{fgvcaircraft}: FGVC-Aircraft is a specialized dataset of aircraft across 100 different aircraft model variants (organized into 30 manufacturers and 70 families), with each image annotated with detailed aircraft type, variant, and manufacturer information.
    \item OxfordIIITPet (Pets) \cite{parkhi12a}: The Oxford-IIIT Pet dataset consists of images of cats and dogs across 37 different pet breeds, with each image annotated with breed labels and pixel-level segmentation masks for pet boundaries and head/body parts.
    \item Country211 \cite{CLIP_country211}: The Country211 dataset contains images of landmarks, objects, and scenes collected from 211 different countries, with roughly 1,000 images per country, designed to test computer vision models' ability to recognize location-specific visual features.
    \item SUN397 \cite{xiao2010}: The SUN397 (Scene UNderstanding) dataset contains 397 diverse scene categories ranging from indoor locations like airports to outdoor scenes like beaches, with at least 100 images per category, designed to evaluate scene recognition capabilities.
    \item DTD \cite{cimpoi14describing}: The Describable Textures Dataset (DTD) consists of 47 different texture categories described by human-chosen adjectives like "cracked," "woven," or "marbled," with images collected from the wild to capture real-world textural variations.
\end{itemize}
For language benchmarks we have:
\begin{itemize}
    \item ToxiGen \cite{toxigen}: ToxiGen is a dataset of approximately 250,000 synthetic toxic and non-toxic text examples generated using large language models and human curation, designed to help test and evaluate content moderation systems for detecting varied forms of harmful content like hate speech and toxicity.
    \item RealToxicityPrompts (RealTox) \cite{realtoxicityprompts}: RealToxicityPrompts is a dataset of approximately 100,000 naturally occurring prompts scraped from the internet and scored for toxicity, designed to test language models' tendency to generate toxic content when given various context prompts.
\end{itemize}

\section{Task arithmetic on noisy dataset using decomposed task vectors}
\label{sec:noisydataset}
For evaluating each component of the decomposed task vectors in the presence of noise, we use the MNIST and MNIST-C datasets \cite{mu2019mnist}. The MNIST-C dataset is a corrupted variant of the MNIST dataset with varying levels of corruption intended for robustness evaluation. We provide the result of task arithmetic using GTSRB and SVHN vectors with MNIST and MNIST-C as the target datasets. This way, we evaluate the impact of the shared and unique component of GTSRB and SVHN on MNIST and its adversarial variant. For results on MNIST-C, we use the glass blur variation of this dataset.

In Figure \ref{fig:mnist}, the addition of SVHN allows for improvement of performance on MNIST while the unique component of this vector also allows for similar improvements. This shows that the two tasks are aligned. On the other hand, when facing a corrupted dataset where task-specific knowledge from other datasets is not expected to be helpful (Figure \ref{fig:mnistc}), the performance from the unique component of SVHN is severely degraded. This shows that only the shared component carried the concepts shared between the two tasks that were helpful for transfer learning. In line with this observation, the shared component consistently improves performance on both noisy and normal datasets. 

\begin{figure}[h]
    \centering
    \begin{subfigure}[b]{0.5\textwidth}
        \centering
        \includegraphics[width=0.99\textwidth]{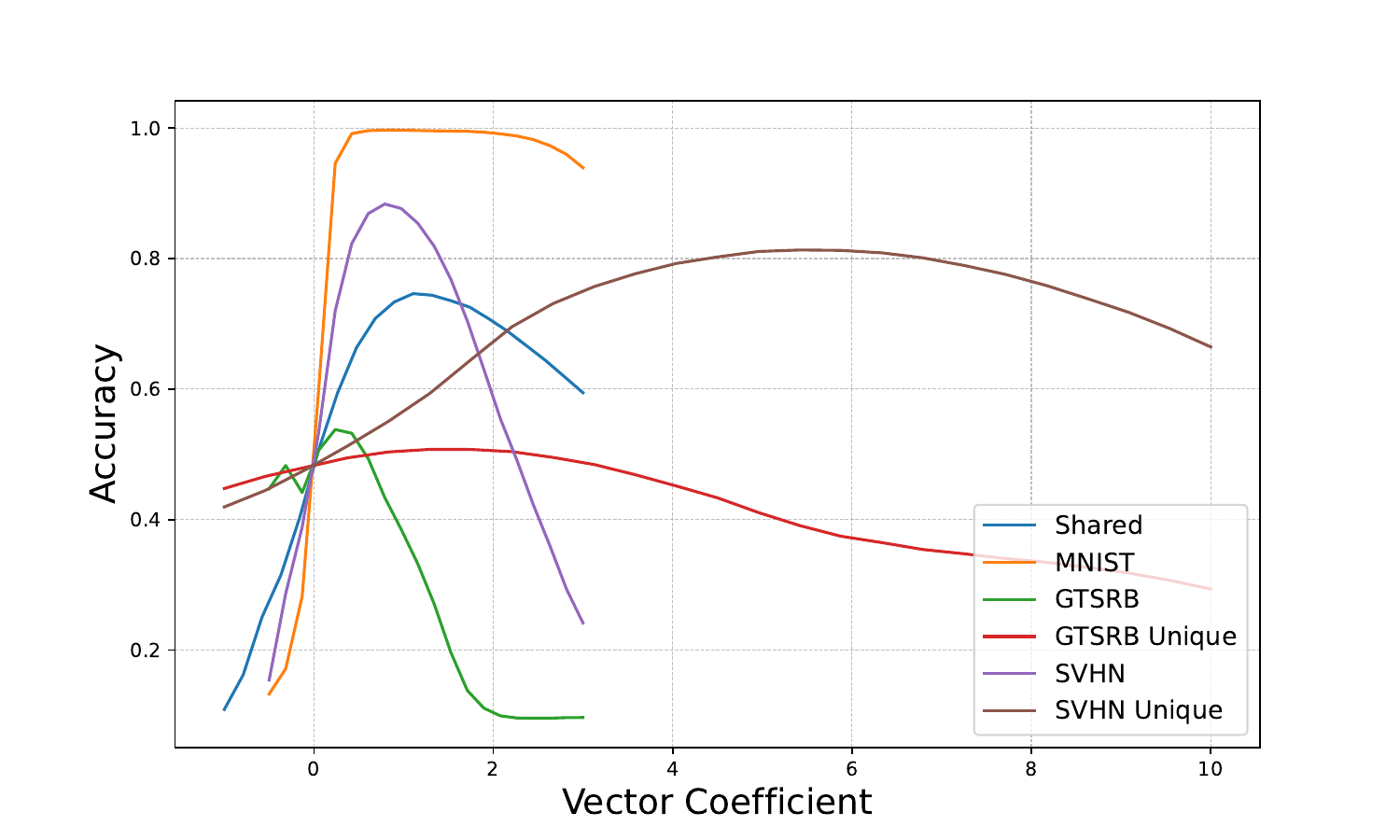}
        \caption{Task arithmetic on MNIST using vectors from GTSRB and SVHN.}
        \label{fig:mnist}
    \end{subfigure}
    \vspace{1em}
    \begin{subfigure}[b]{0.5\textwidth}
        \centering
        \includegraphics[width=0.99\textwidth]{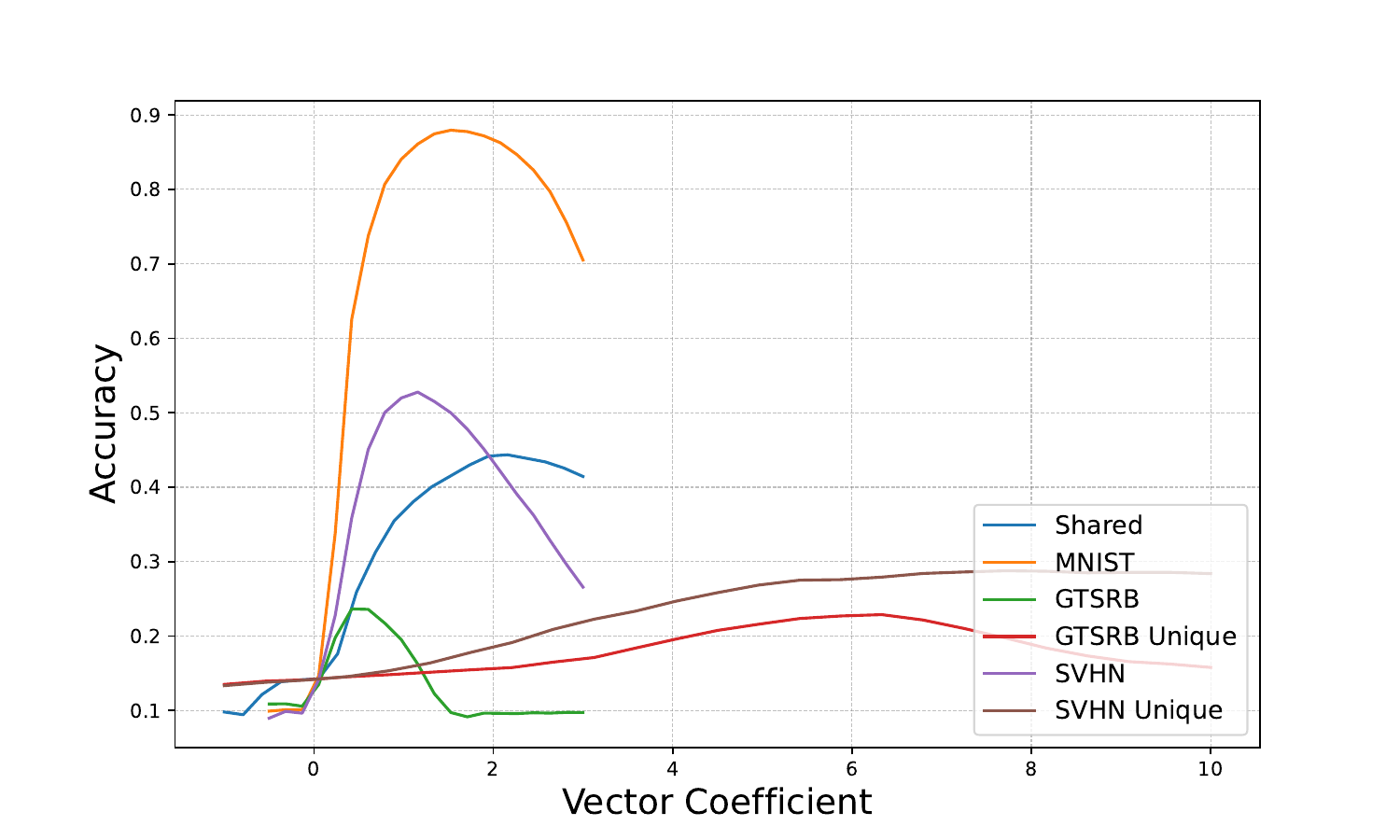}
        \caption{Task arithmetic on MNIST-C using vectors from GTSRB and SVHN.}
        \label{fig:mnistc}
    \end{subfigure}
    \caption{Task arithmetic on MNIST shows that SVHN allows for better performance where GTSRB has negligible effects. This shows that the requirement of digit recognition from SVHN allows for appropriate transfer learning to MNIST. This is while the shared component between GTSRB and SVHN is able to capture the digit recognition capabilities. MNIST-C still shows improvement with the shared component whereas the unique vectors from SVHN and GTSRB fail since these components embed the task-specific knowledge that fails to deliver on corrupted MNIST variant.}
    \label{fig:mnists}
\end{figure}

\section{Images generated by each style}
\label{sec:eachstyle}
Here we provide the outputs of the individual fine-tuned checkpoints used for image generation in Section \ref{sec:imagegeneration} to visualize the style associated with each fine-tuned checkpoint. We use 3 fine-tuned checkpoints of Flux Schnell 1.0 \cite{FLUX1_schnell} trained on anime style images \cite{civitai_flat_colour_anime,civitai_flux_ghibli,civitai_model640126}, 1 checkpoint for black and white images \cite{civitai_1930s_photography_flux} and 1 checkpoint for pixel art style images \cite{civitai_pokemon_trainer}. An example of each style is shown in Figure \ref{fig:generated1} to Figure \ref{fig:generated5}.

\begin{figure}[h]
    \centering
    \includegraphics[width=0.47\textwidth]{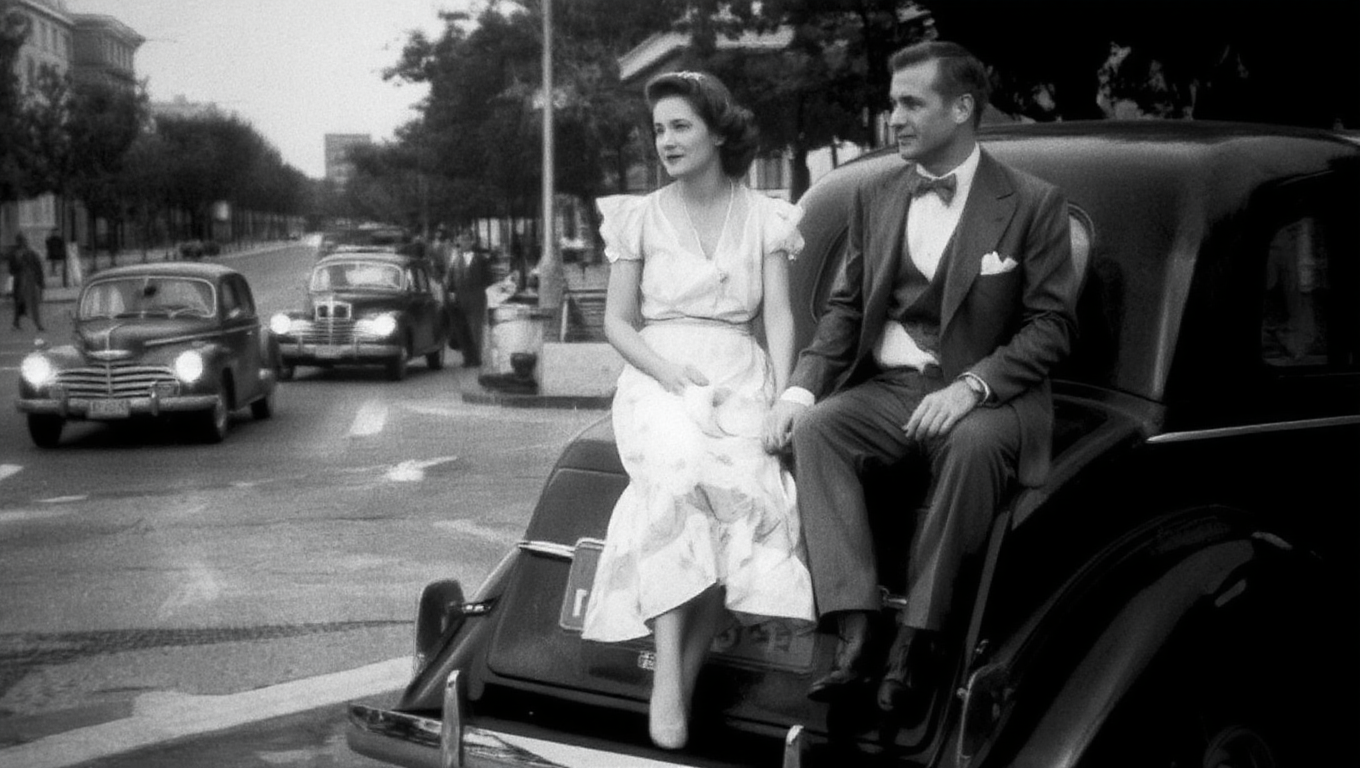}
    \caption{Output of fine-tuned checkpoint with black and white photography style. Prompt: "1930s photography mode, monochrome, a couple sits on the trunk of a vintage car, their hands lightly touching as traffic passes by"}
    \label{fig:generated1}
\end{figure}

\begin{figure}[h]
    \centering
    \includegraphics[width=0.47\textwidth]{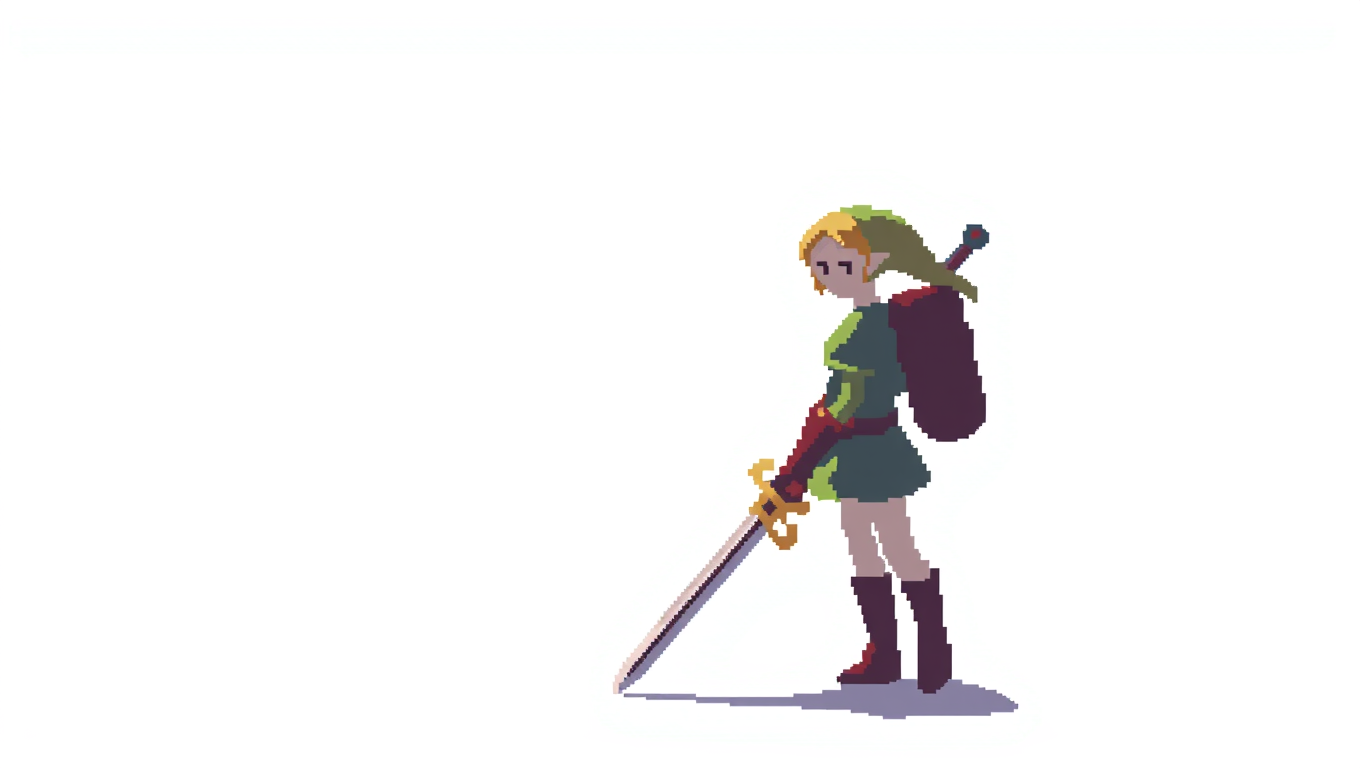}
    \caption{Output of fine-tuned checkpoint with pixel art style. Prompt: "link of zelda, with a sword simple background"}
    \label{fig:generated2}
\end{figure}

\begin{figure}[h]
    \centering
    \includegraphics[width=0.47\textwidth]{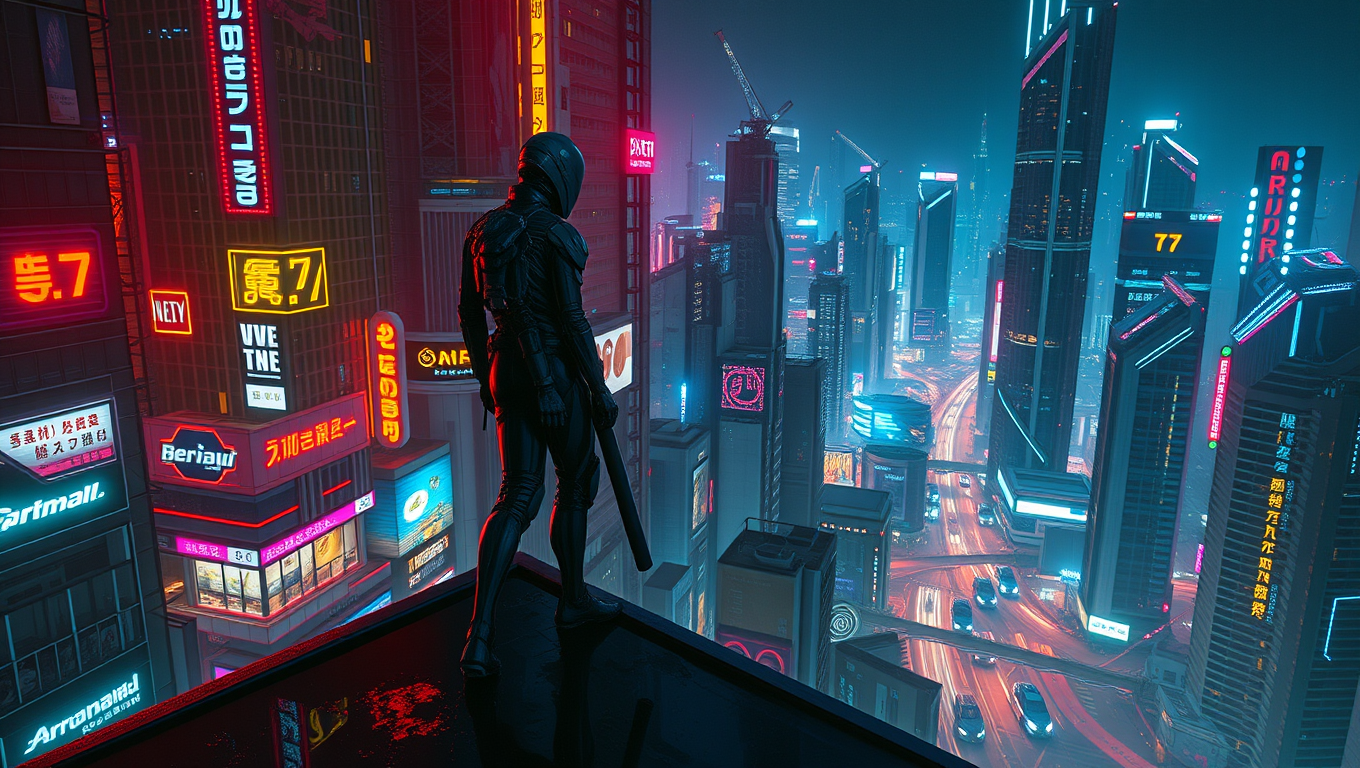}
    \caption{Output of fine-tuned checkpoint with generic anime style. Prompt: "Cyberpunk megacity at midnight, neon-soaked streets reflecting in puddles. A lone figure in a high-tech suit stands atop a skyscraper, city sprawling below. Harsh contrasts, vibrant colors, detailed mechanical elements"}
    \label{fig:generated3}
\end{figure}

\begin{figure}[h]
    \centering
    \includegraphics[width=0.47\textwidth]{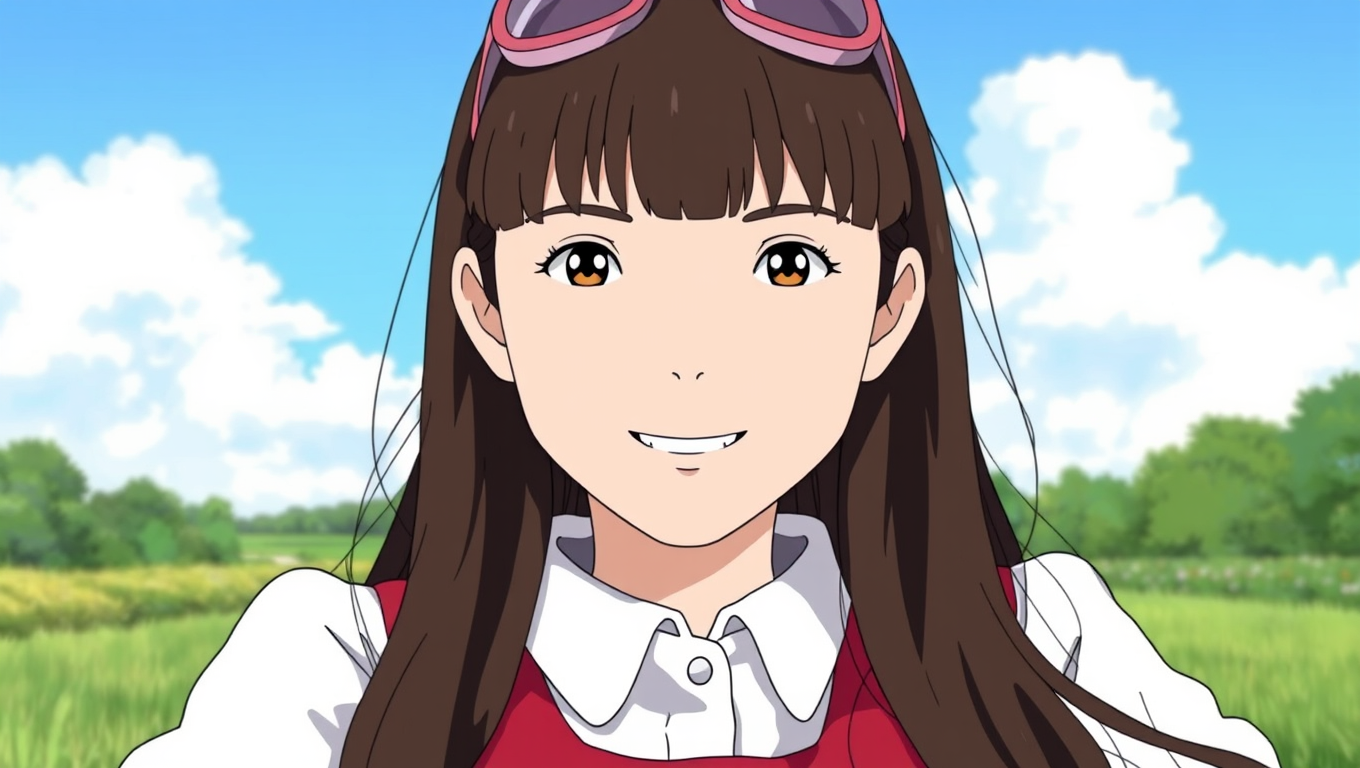}
    \caption{Output of fine-tuned checkpoint on Studio Ghibli style anime. Prompt: "smile, dress, bangs, red dress, goggles on head, day, solo, white shirt, long hair"}
    \label{fig:generated4}
\end{figure}

\begin{figure}[h]
    \centering
    \includegraphics[width=0.47\textwidth]{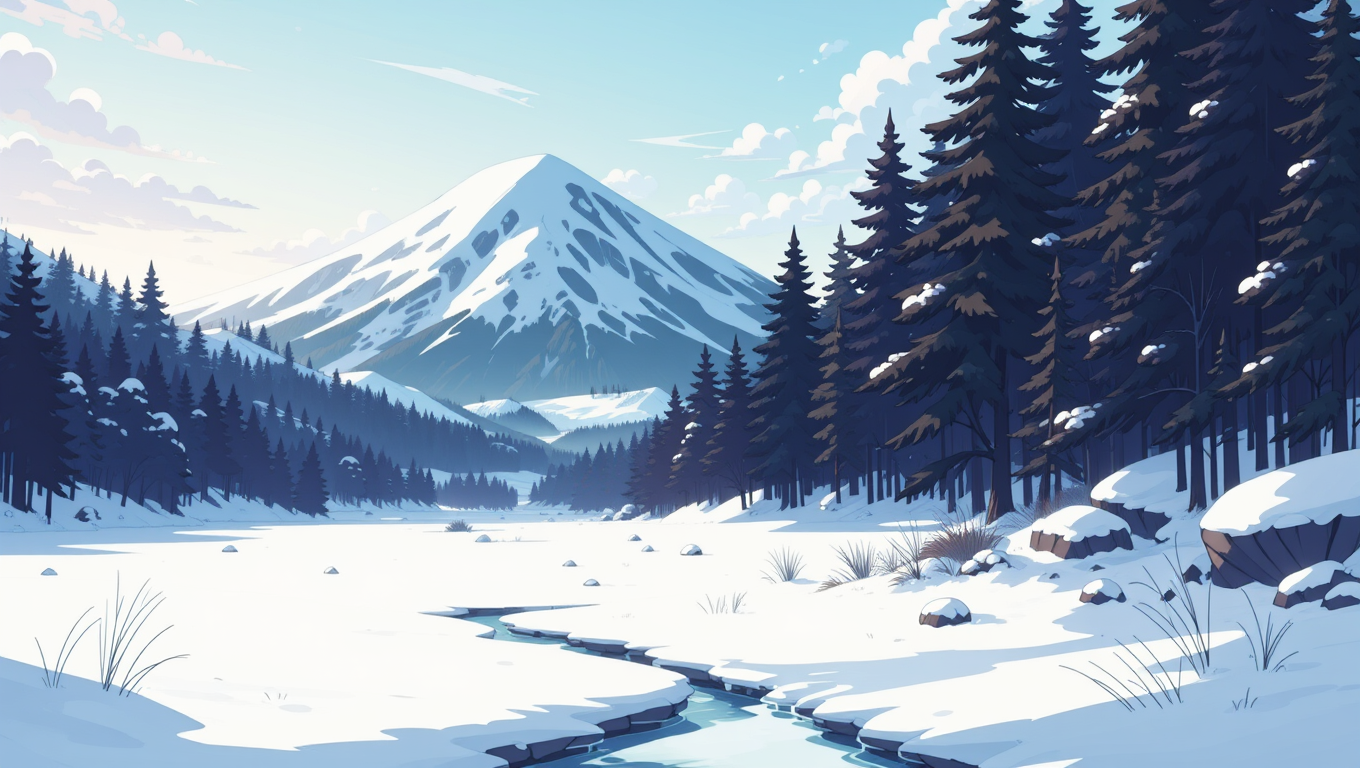}
    \caption{Output of fine-tuned checkpoint on flat color anime style. Prompt: "Flat colour anime style image showing a snowy landscape with a mountain in the distance"}
    \label{fig:generated5}
\end{figure}

\section{Style mixing using unique components}
\label{sec:morestylemixing}
In this section, we provide the results of style mixing using 2 and 3 styles. The results in Figure \ref{fig:texttoimage} showed that when mixing 4 styles, the shared vector component aggressively pushes the model toward concepts shared between the styles leading to degraded outputs and underemphasis of unique concepts embedded in each style. Note that for every style mixing experiment, we start with the vector coefficients suggested by the fine-tuned checkpoint's author and increase this coefficient for all styles simultaneously for all vectors for both our approach and the baseline until we obtain the mixed style.

Figure \ref{fig:texttoimage2} shows the results of mixing 3 different anime styles across 3 different seeds. Similar results to Figure \ref{fig:texttoimage} are observed where multiple similar styles cause degradation in the output of the text-to-image model whereas mixing the unique components allows for outputs with the desired mixing. Even with 2 styles, it is still possible to get degraded outputs as shown in Figure \ref{fig:texttoimage3} where a generic anime style is mixed with Studio Ghibli anime style. The unique components, on the other hand, are still able to adequately mix the styles regardless of the number of styles in the pool.

The reported visualizations so far have been using styles that have potential conflicts in their shared components since multiple anime styles were used. We also provide outputs of the same text-to-image model when using a black-and-white style alongside a pixel art style. Compared to the previous experiments, these two vectors are not expected to have major conflicts in the specific intended style by the fine-tuned checkpoints. These results are provided in Figure \ref{fig:texttoimage3}. While our style mixing approach with unique vectors adheres to both the given prompt and the desired style, the vanilla mixing method fails to incorporate either the pixel art style or the black and white style in different seeds. 

\begin{figure*}[bht]
    \begin{subfigure}[b]{0.45\textwidth}
        \centering
        \includegraphics[width=\textwidth]{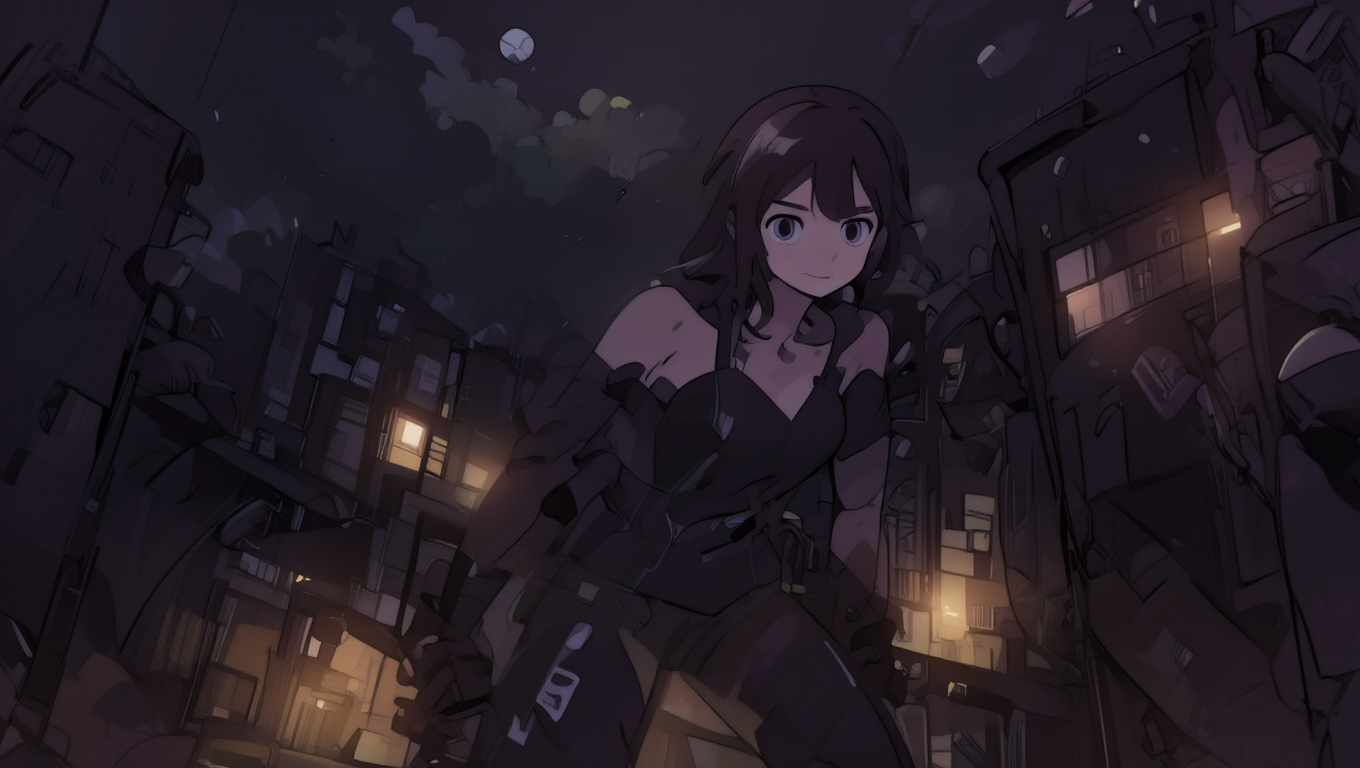}
        \label{fig:texttoimage2_1}
    \end{subfigure}
    \hfill
    \begin{subfigure}[b]{0.45\textwidth}
        \centering
        \includegraphics[width=\textwidth]{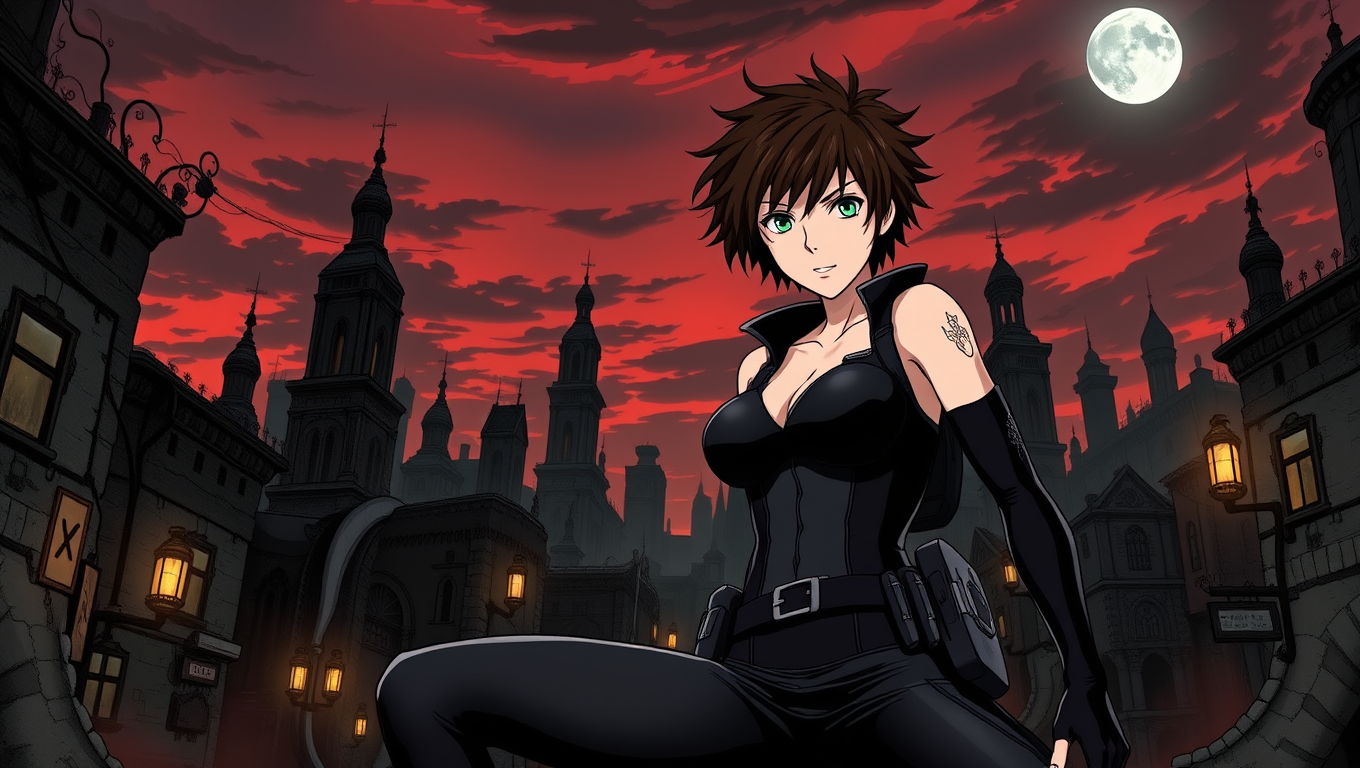}
        \label{fig:texttoimage2_2}
    \end{subfigure}
    \vfill
    \begin{subfigure}[b]{0.45\textwidth}
        \centering
        \includegraphics[width=\textwidth]{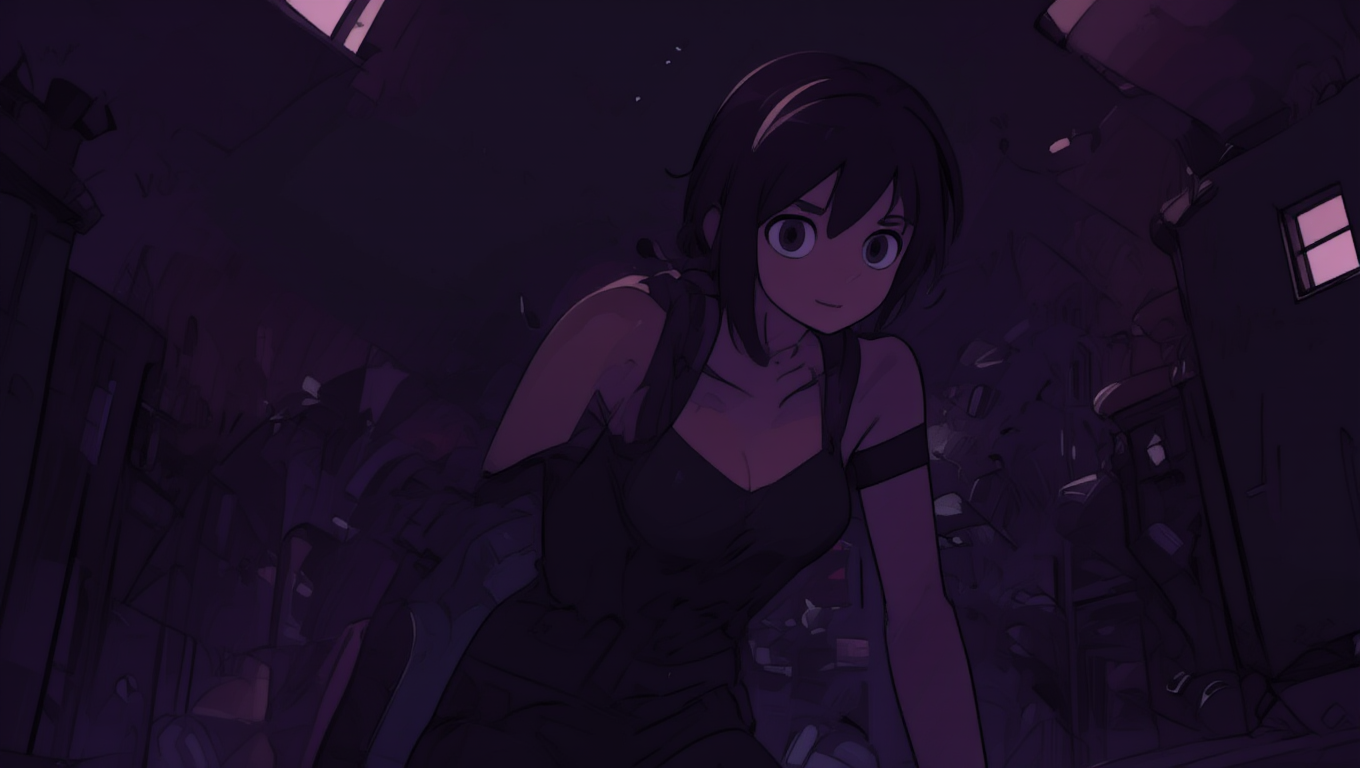}
        \label{fig:texttoimage2_3}
    \end{subfigure}
    \hfill
    \begin{subfigure}[b]{0.45\textwidth}
        \centering
        \includegraphics[width=\textwidth]{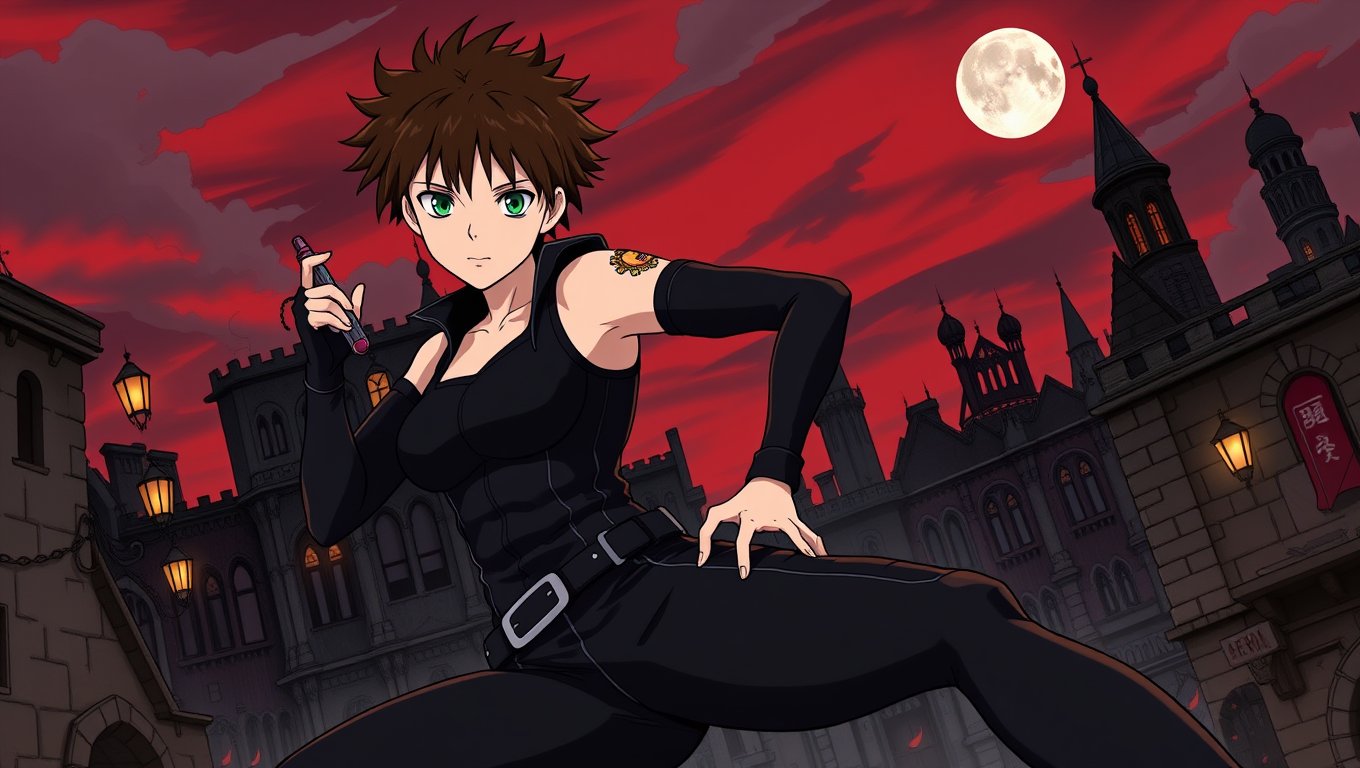}
        \label{fig:texttoimage2_4}
    \end{subfigure}
    \vfill
    \begin{subfigure}[b]{0.45\textwidth}
        \centering
        \includegraphics[width=\textwidth]{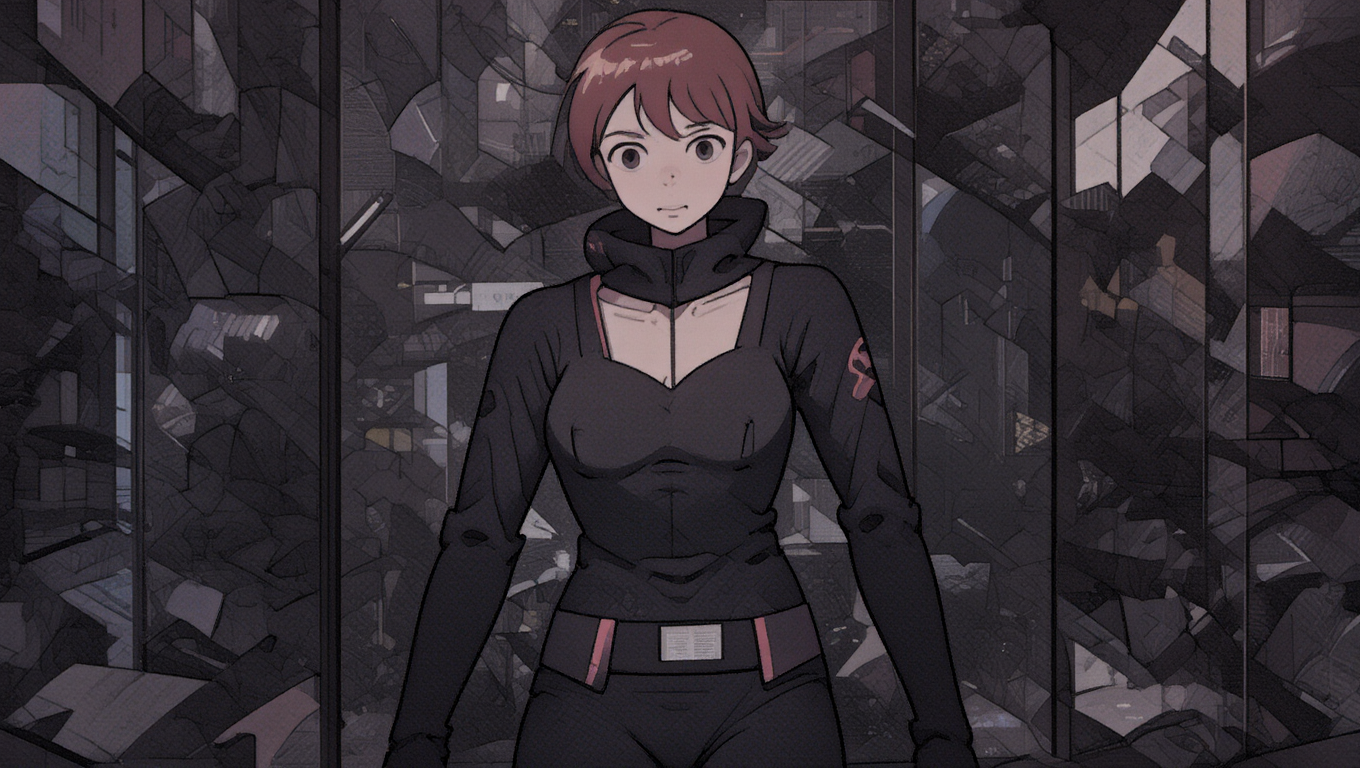}
        \caption{Vanilla task vectors}
        \label{fig:texttoimage2_5}
    \end{subfigure}
    \hfill
    \begin{subfigure}[b]{0.45\textwidth}
        \centering
        \includegraphics[width=\textwidth]{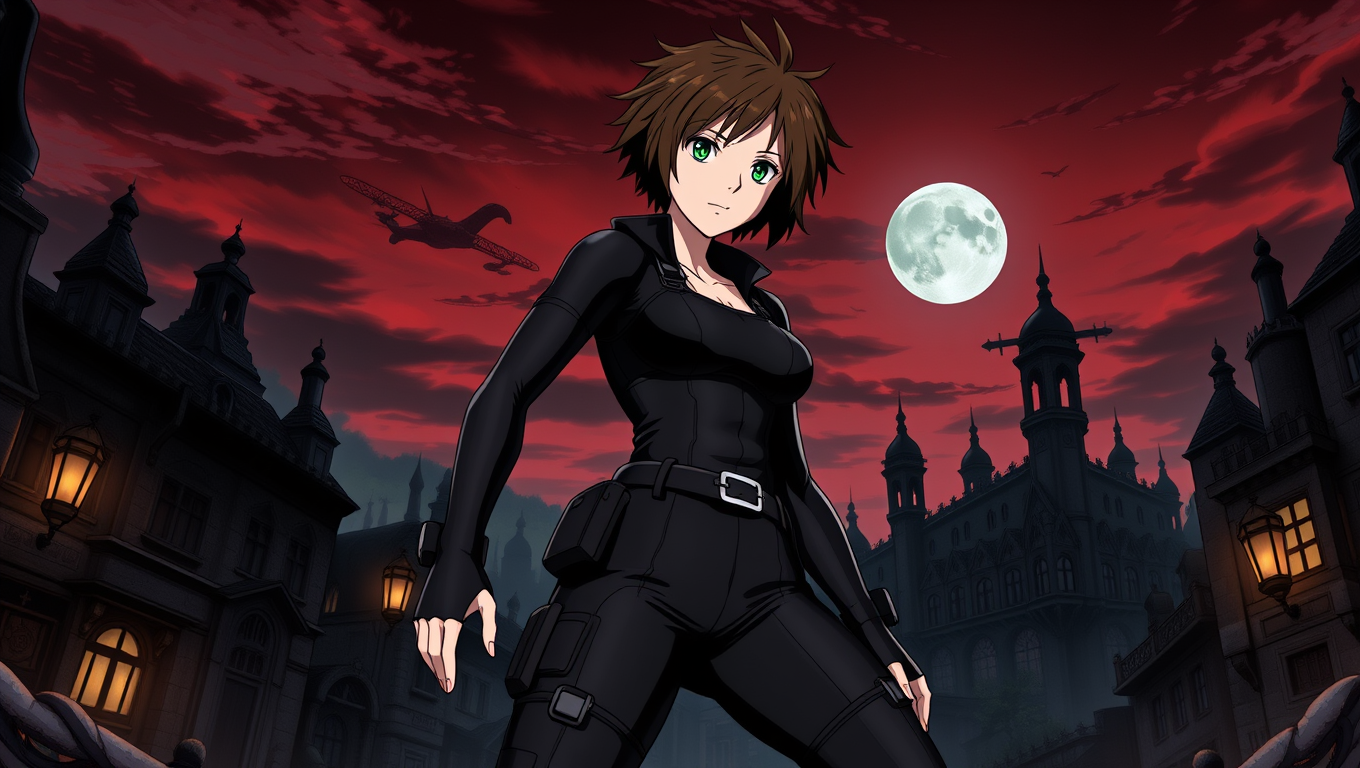}
        \caption{Unique vectors (ours)}
        \label{fig:texttoimage2_6}
    \end{subfigure}
    \caption{Outputs of a text-to-image model mixing 3 anime styles across 3 different seeds with the prompt "anime style image showing an athletic adult woman dressed in a sleek black and dark grey jumpsuit, with a utility belt around her waist, armed with an arsenal of vampire-hunting gadgets."}
    \label{fig:texttoimage2}
\end{figure*}

\begin{figure*}[bht]
    \centering
    \begin{subfigure}[b]{0.45\textwidth}
        \centering
        \includegraphics[width=\textwidth]{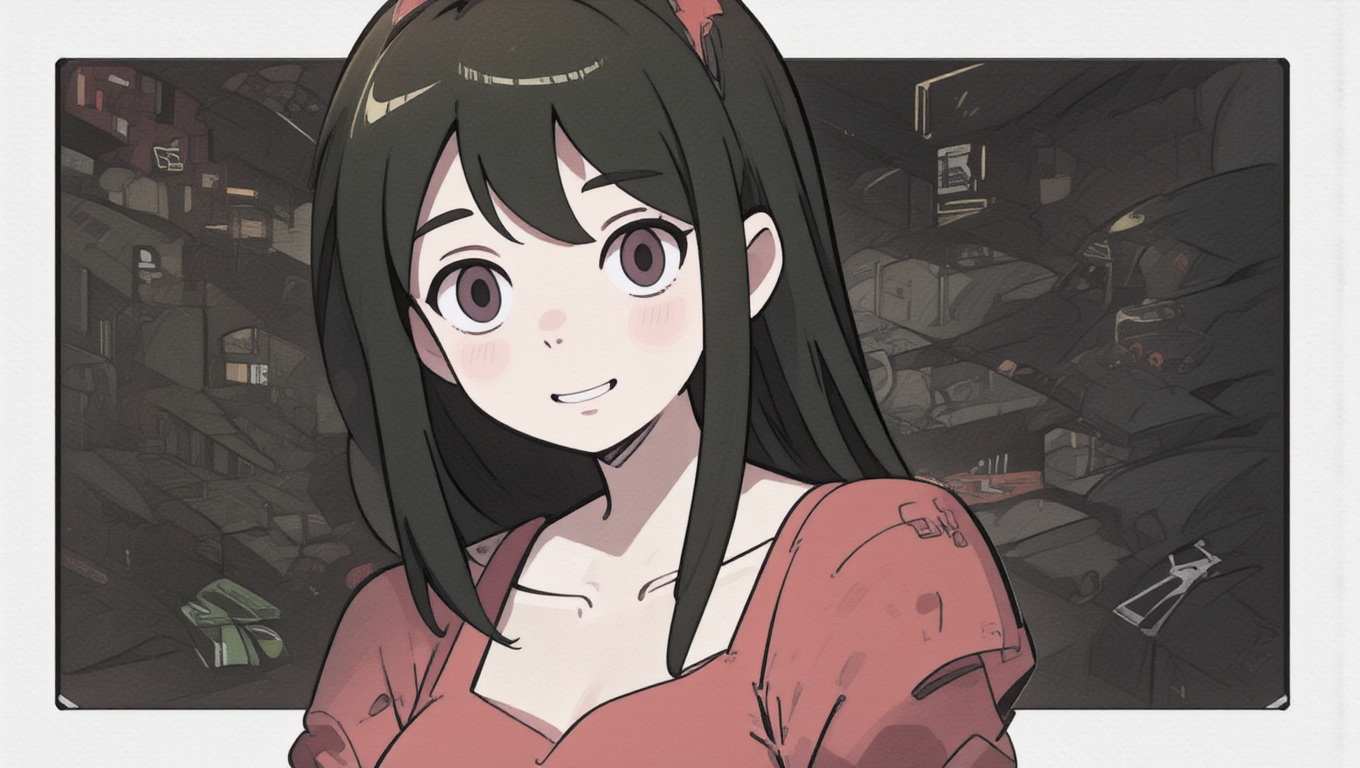}
        \label{fig:texttoimage3_1}
    \end{subfigure}
    \hfill
    \begin{subfigure}[b]{0.45\textwidth}
        \centering
        \includegraphics[width=\textwidth]{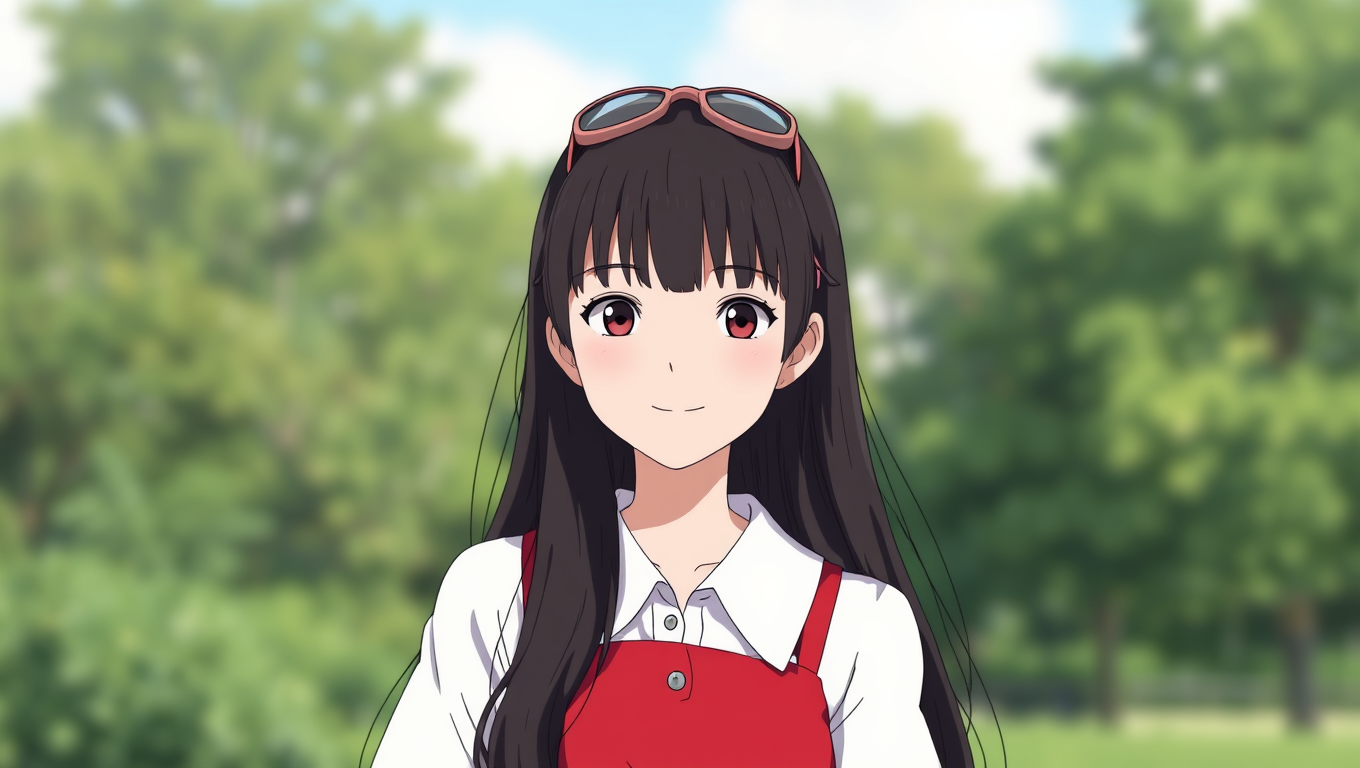}
        \label{fig:texttoimage3_2}
    \end{subfigure}
    \caption{Outputs of a text-to-image model mixing 2 anime styles (generic anime style and Studio Ghibli style) with the prompt "smile, dress, bangs, red dress, goggles on head, day, solo, white shirt, long hair"}
    \label{fig:texttoimage3}
\end{figure*}

\begin{figure*}[bht]
    \begin{subfigure}[b]{0.45\textwidth}
        \centering
        \includegraphics[width=\textwidth]{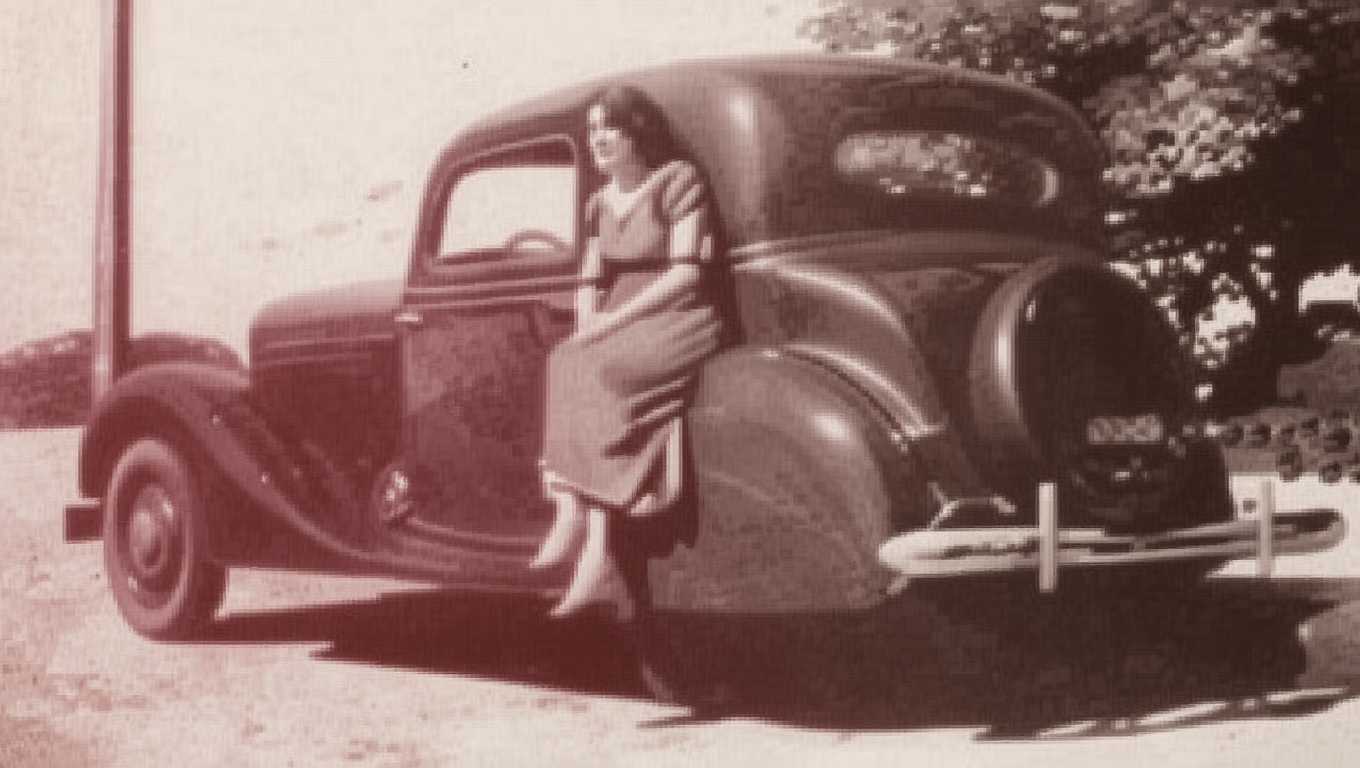}
        \label{fig:texttoimage3_1}
    \end{subfigure}
    \hfill
    \begin{subfigure}[b]{0.45\textwidth}
        \centering
        \includegraphics[width=\textwidth]{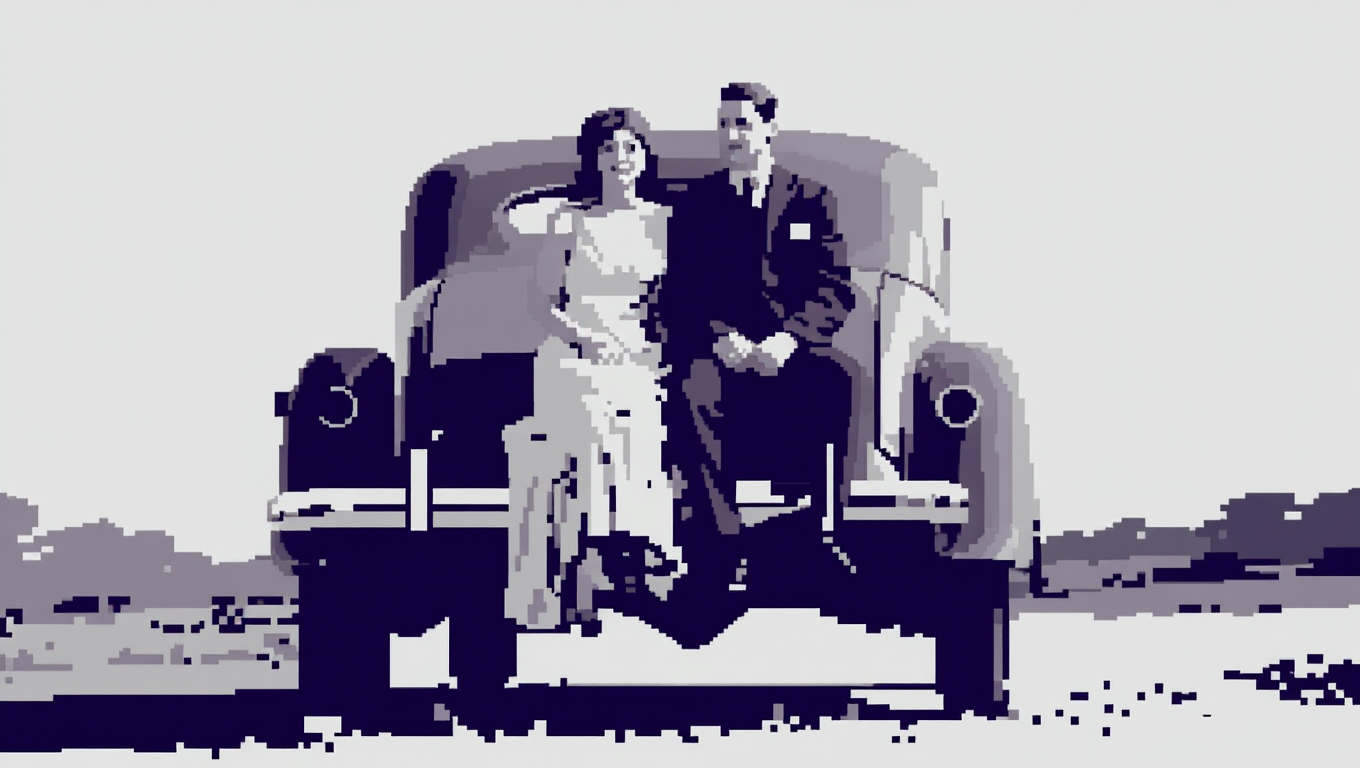}
        \label{fig:texttoimage3_2}
    \end{subfigure}
    \vfill
    \begin{subfigure}[b]{0.45\textwidth}
        \centering
        \includegraphics[width=\textwidth]{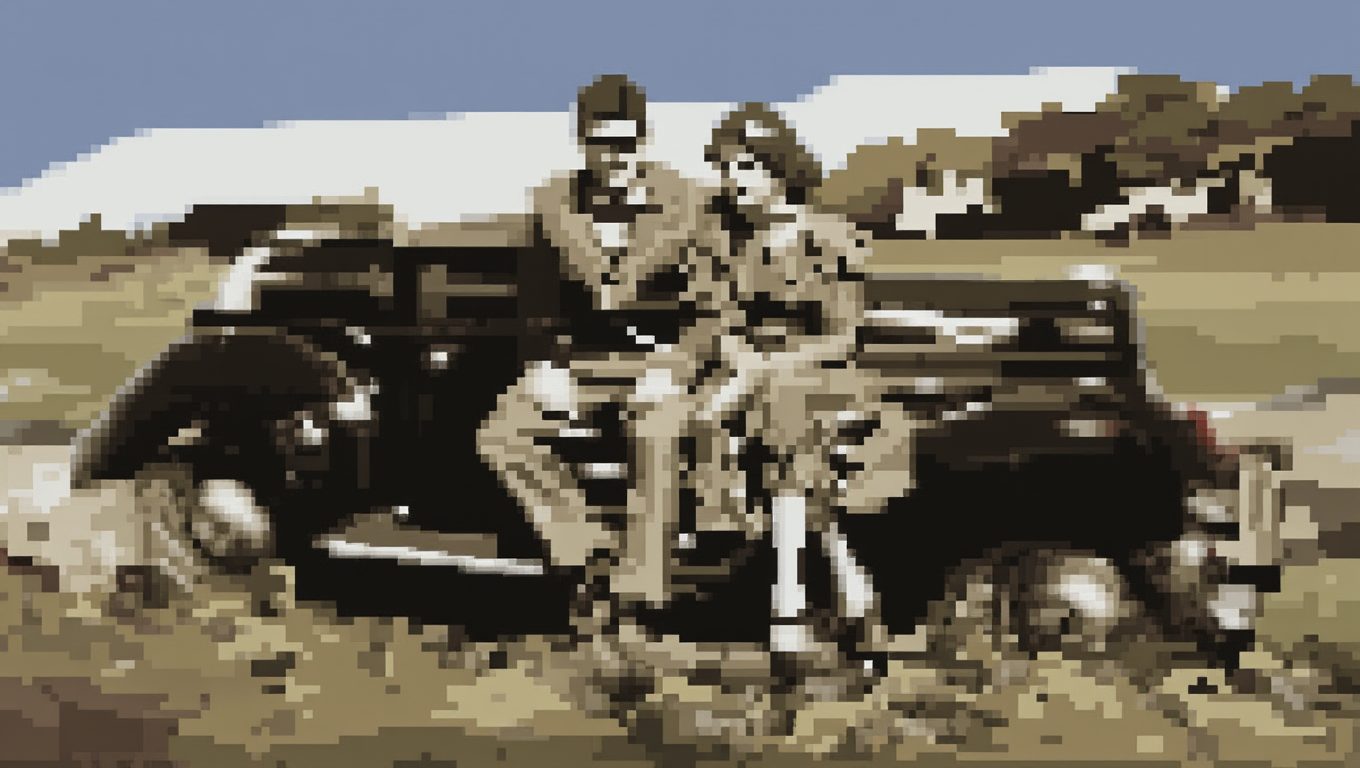}
        \label{fig:texttoimage3_3}
    \end{subfigure}
    \hfill
    \begin{subfigure}[b]{0.45\textwidth}
        \centering
        \includegraphics[width=\textwidth]{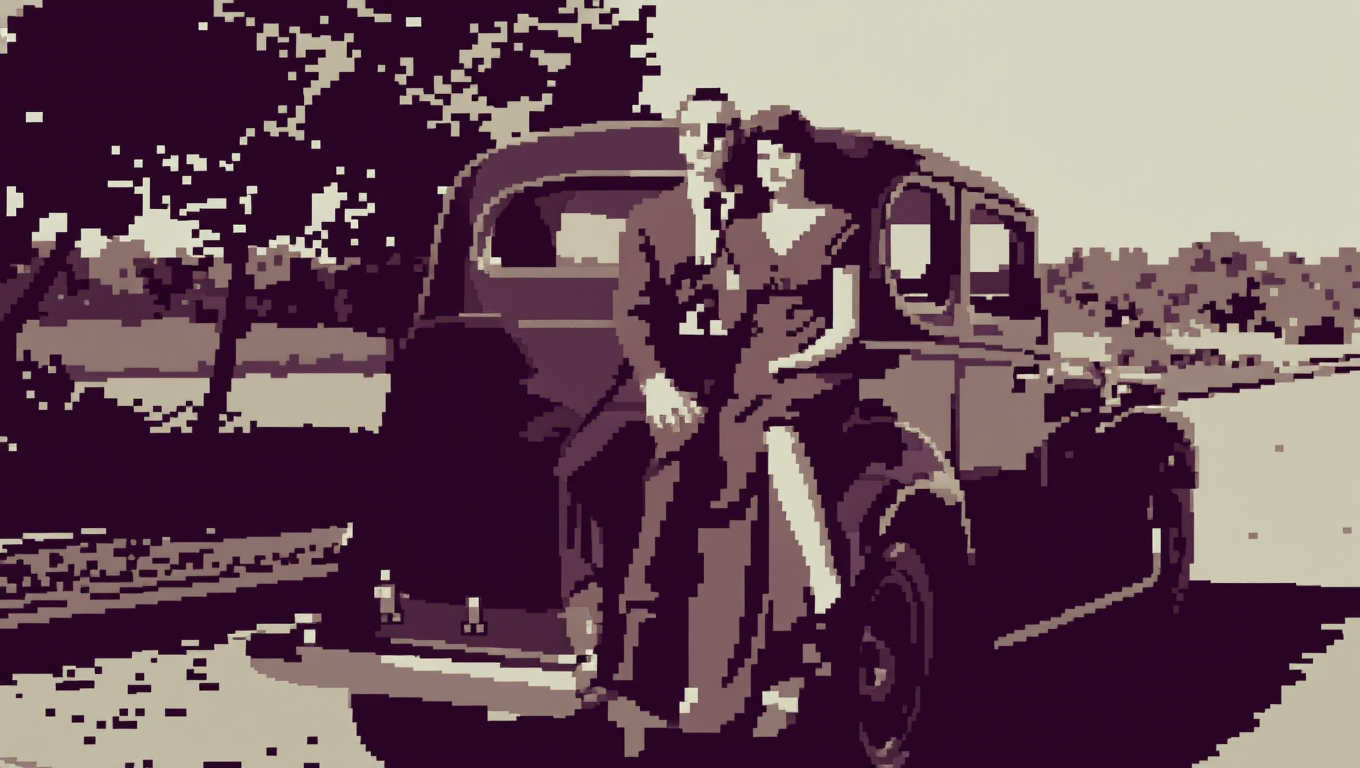}
        \label{fig:texttoimage3_4}
    \end{subfigure}
    \vfill
    \begin{subfigure}[b]{0.45\textwidth}
        \centering
        \includegraphics[width=\textwidth]{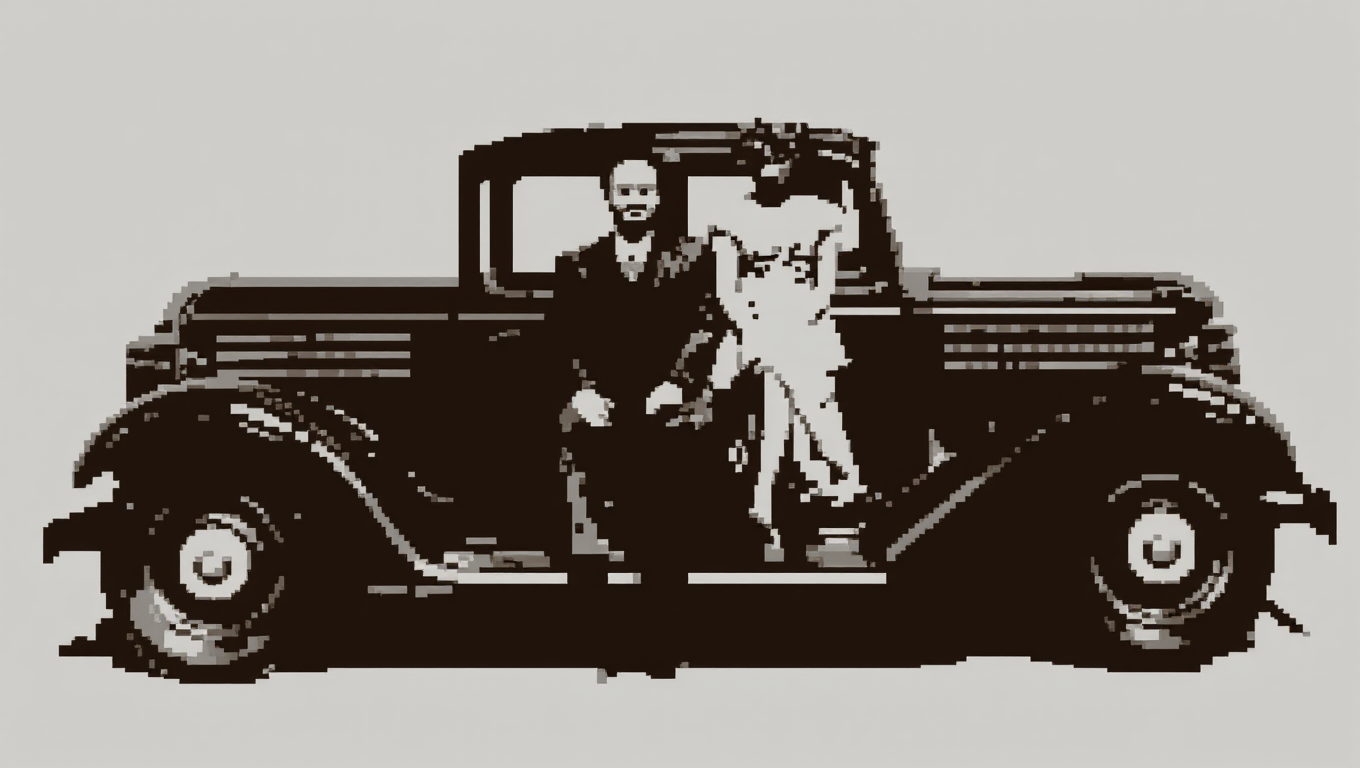}
        \caption{Vanilla task vectors}
        \label{fig:texttoimage3_5}
    \end{subfigure}
    \hfill
    \begin{subfigure}[b]{0.45\textwidth}
        \centering
        \includegraphics[width=\textwidth]{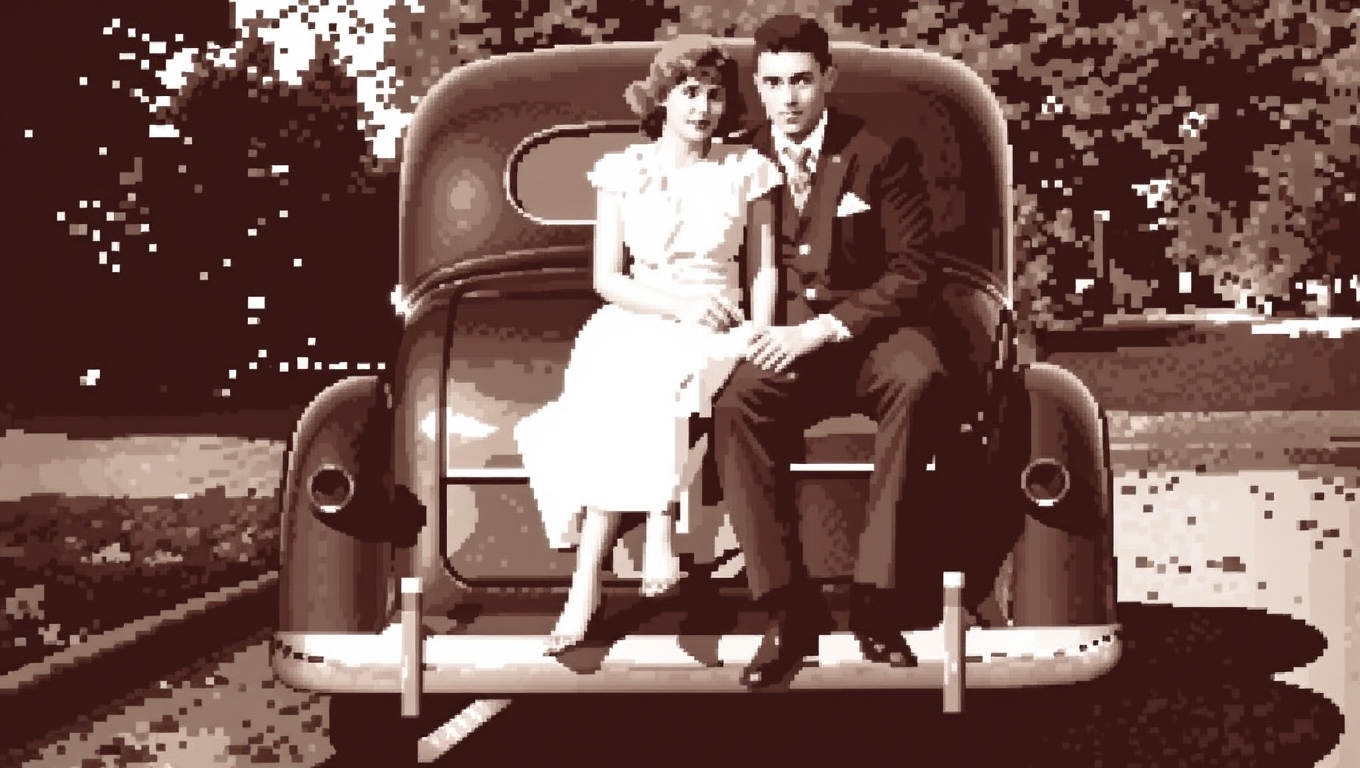}
        \caption{Unique vectors (ours)}
        \label{fig:texttoimage3_6}
    \end{subfigure}
    \caption{Outputs of a text-to-image model mixing black-and-white style with pixel art style across 3 different seeds with the prompt "1930s photography mode, monochrome, a couple sits on the trunk of a vintage car, their hands lightly touching as traffic passes by."}
    \label{fig:texttoimage3}
\end{figure*}

\section{Implementation details for LLAMA2 finetuning}
\label{sec:llamafinetune}

Here we provide the implementation details on finetuning LLAMA2 \cite{touvron2023llama2openfoundation} on the RealToxicityPrompts and ToxiGen datasets. We fine-tuned the LLaMA-2-7B model using Parameter-Efficient Fine-Tuning (PEFT) with Weight-Decomposed Low-Rank Adaptation (DoRA) \cite{liu2024dora}. The model was trained on toxic content from either the ToxiGen dataset or the Real Toxicity Prompts dataset, filtering for examples with toxicity scores above 0.5 or binary toxic labels respectively.
For the DoRA configuration, we used a rank of 256 (r=256) and scaling factor 512, with a dropout rate of 0.02. We applied LoRA to all linear layers in the model for improved adaptation. The model was trained using mixed-precision (FP16) with gradient checkpointing enabled to optimize memory usage. The training process utilized the following hyperparameters:

\begin{itemize}
    \item Learning rate: 2e-4
    \item Batch size: 256 (without gradient accumulation)
    \item Number of epochs: 7 for RealToxicityPrompts and 15 for ToxiGen
    \item Optimizer: ADAMW \cite{loshchilov2018decoupled} with weight decay of 0.01
\end{itemize}

We preprocessed the data by concatenating prompts and continuations, using special token masking to ensure the model only learns from the continuation portion during training. 

\section{Checkpoints in pool of LLAMA2 task vectors}
\label{sec:llamacheckpoints}
The checkpoints available in the pool of task vectors for the toxicity removal experiment in Section \ref{sec:toxicityremoval} are listed as follows:
\begin{itemize}
    \item Odia LLAMA-2-7B \cite{odia_llama2_7b_base}: Instruction fine-tuned model with 180000 instructions focused on question answering and content generation.
    \item Dolphin LLAMA2-7B \cite{dolphin_llama2_7b}: Uncensored instruction fine-tuned model trained on FLANv2 instructions with GPT-augmented results.
    \item EMMA-500 LLAMA2-7B \cite{emma_500_llama2_7b}: Finetuned LLAMA2 checkpoint with Multilingual training dataset spanning 500 languages and 74 billion tokens.
    \item Ultrachat 200k LLAMA2-7B \cite{llama2_ultrachat_syn200k}: Finetuned LLAMA2 model on a heavily filtered version of the ultrachat dataset focusing on multi-round dialogue data.
    \item Ultrafeekback \cite{llama2_sft_full}: Finetuned LLAMA2 model on ultrafeedback preference dataset.
    \item LLAMA2-Chat \cite{llama2_7b_chat_hf}: Original chat model released by Meta as a dialogue model.
    \item LLAMA2-Chat-Uncensored \cite{llama2_7b_chat_uncensored}: Uncensored variant of the LLAMA2-Chat model trained on the Vicuna conversation dataset.
    \item Physics LLAMA2-7B \cite{llama2_7b_physics}: Finetuned LLAMA2 model using a dataset consisting of 20K problem-solution pairs generated using GPT4.
    \item Asclepius LLAMA2-7B \cite{asclepius_llama2_7b}: Finetuned LLAMA2 model using a dataset of clinical notes.
    \item Roman Empire LLAMA2-7B \cite{llama2_7b_roman_empire_qa}: Finetuned LLAMA2 model using questions and answers about the Roman Empire.
\end{itemize}

The following details the naming used for task vectors pools in \cref{fig:negation}:
\begin{itemize}
    \item LLAMA-Chat: The original chat variant of the LLAMA2-7b model released by Meta.
    \item All-LLAMA-Chats: The original LLAMA2 chat model alongside the uncensored version termed LLAMA2-Chat-Uncensored \cite{llama2_7b_chat_uncensored}.
    \item Narrow: The combination of above task vectors with Roman Empire LLAMA2-7B, Asclepius LLAMA2-7B, Physics LLAMA2-7B in the pool.
    \item Small Instruct: Combination of Odia LLAMA-2-7B, Dolphin LLAMA2-7B, LLAMA2-Chat, and LLAMA2-Chat-Uncensored.
    \item Random: Randomly chosen combination of Roman Empire LLAMA2-7B, EMMA-500 LLAMA2-7B, LLAMA2-Chat, Ultrachat 200k LLAMA2-7B, and Physics LLAMA2-7B.
    \item Instruct EMMA: Combiation of Odia LLAMA-2-7B, Dolphin LLAMA2-7B, Ultrachat 200k LLAMA2-7B, Ultrafeekback, LLAMA2-Chat, LLAMA2-Chat-Uncensored, and EMMA-500 LLAMA2-7B.
    \item All-Models: All the task vectors detailed above.

\end{itemize}

\end{document}